\documentclass[10pt,twocolumn,letterpaper]{article}

\usepackage{cvpr}              

\usepackage{graphicx}
\usepackage{amsmath}
\usepackage{amssymb}
\usepackage{booktabs}
\usepackage[dvipsnames]{xcolor}
\usepackage{makecell}
\usepackage{pifont}
\usepackage[normalem]{ulem}
\usepackage{multirow}
\usepackage{float}

\usepackage{placeins} 

\usepackage{bbding}
\newcommand{\snow}{{\SnowflakeChevron}} 

\usepackage[pagebackref,breaklinks,colorlinks]{hyperref}

\definecolor{forest}{RGB}{85,91,25}
\definecolor{countryside}{RGB}{111,163,27}
\definecolor{rural_farmland}{RGB}{237,197,0}
\definecolor{highway}{RGB}{105,110,106}
\definecolor{low_density_residential}{RGB}{13,213,148}
\definecolor{community_buildings}{RGB}{0,147,230}
\definecolor{high_density_residential}{RGB}{213,42,0}

\definecolor{General}{RGB}{1,115,178} 
\definecolor{Cell}{RGB}{2, 158, 115} 
\definecolor{Scratch}{RGB}{222, 143, 5} 
\usepackage[capitalize]{cleveref}
\crefname{section}{Sec.}{Secs.}
\Crefname{section}{Section}{Sections}
\Crefname{table}{Table}{Tables}
\crefname{table}{Tab.}{Tabs.}

\newcommand\blfootnote[1]{%
  \begingroup
  \renewcommand\thefootnote{}\footnote{#1}%
  \addtocounter{footnote}{-1}%
  \endgroup
}

\global\long\def\allAgents{\mathbf{A}}%
\global\long\def\allModels{\mathbf{F}}%
\global\long\def\allDistributions{\mathbf{P}}
\global\long\def\allDatastreams{\mathbf{X}}
\global\long\def\allEnvironments{\mathbf{E}}%
\global\long\def\task{\mathcal{\tau}}%
\global\long\def\datastream#1#2{\mathcal{X}_{#1}^{#2}}%
\global\long\def\replayBuffer#1#2{\mathcal{R}_{#1}^{#2}}

\global\long\def\loss{\mathcal{L}}
\global\long\def\distance{L}
\global\long\def\replayIndex{i}
\global\long\def\replayBufferSize{R}
\global\long\def\distribution#1#2{\mathcal{P}_{#1}^{#2}}%
\global\long\def\time{t}%
\global\long\def\lastTime{T}%
\global\long\def\agent#1{a_{#1}}%
\global\long\def\model#1#2{f_{#1}^{#2}}%
\global\long\def\sample#1#2{x_{#1}^{#2}}%
\global\long\def\prediction#1#2{\hat{y}_{#1}^{#2}}%
\global\long\def\pseudo#1#2{\tilde{y}_{#1}^{#2}}%
\global\long\def\environment#1#2{e_{#1}^{#2}}%
\global\long\def\environmentAlone#1#2{{#1}^{#2}}%
\global\long\def\student#1#2{\mathcal{S}_{#1}^{#2}}%
\global\long\def\teacher#1#2{\mathcal{T}_{#1}^{#2}}%
\global\long\def\agentIndex{n}%
\global\long\def\batchsize{b}%
\global\long\def\framerate#1{r_{#1}}%
\global\long\def\batch#1#2{\mathcal{B}_{#1}^{#2}}%
\global\long\def\agentIndexb{m}%
\global\long\def\environmentIndex{c}%
\global\long\def\agentSize{N}%
\global\long\def\environmentSize{C}%

\def\cvprPaperID{}
\def\confName{}
\def\confYear{}
\newcommand{\mysection}[1]{\vspace{2pt}\noindent\textbf{#1}}

\begin{document}

\title{Multi-Stream Cellular Test-Time Adaptation\\of Real-Time Models Evolving in Dynamic Environments}

\author{Benoît Gérin$^{1,*}$ 
\quad
Anaïs Halin$^{2,*}$  
\quad
Anthony Cioppa$^{2,3,*}$  
\quad
Maxim Henry$^2$  
\and 
Bernard Ghanem$^3$  
\quad
Benoît Macq$^1$ 
\quad
Christophe De Vleeschouwer$^1$ 
\quad
Marc Van Droogenbroeck$^2$  
\and 
$^1$ {\small UCLouvain} \quad $^2$ {\small ULiège} \quad $^3$ {\small KAUST}
\and
{\tt\small benoit.gerin@uclouvain.be, anais.halin@uliege.be, anthony.cioppa@uliege.be}
}
\maketitle

\begin{abstract}
In the era of the Internet of Things (IoT), objects connect through a dynamic network, empowered by technologies like 5G, enabling real-time data sharing.
However, smart objects, notably autonomous vehicles, face challenges in critical local computations due to limited resources. Lightweight AI models offer a solution but struggle with diverse data distributions.
To address this limitation, we propose a novel \emph{Multi-Stream Cellular Test-Time Adaptation} (MSC-TTA) setup where models adapt on the fly to a dynamic environment divided into cells. 
Then, we propose a real-time adaptive student-teacher method that leverages the multiple streams available in each cell to quickly adapt to changing data distributions. 
We validate our methodology in the context of autonomous vehicles navigating across cells defined based on location and weather conditions. To facilitate future benchmarking, we release a new multi-stream large-scale synthetic semantic segmentation dataset, called \emph{DADE}, and show that our multi-stream approach outperforms a single-stream baseline. 
We believe that our work will open research opportunities in the IoT and 5G eras, offering solutions for real-time model adaptation.
\blfootnote{\textbf{(*)} Equal contributions. 
Code and data available at \href{https://github.com/ULiege-driving/MSC-TTA}{github.com/\\ULiege-driving/MSC-TTA} and \href{https://github.com/ULiege-driving/DADE}{github.com/ULiege-driving/DADE}.
}
\end{abstract}

\section{Introduction}\label{sec:introduction}

In the contemporary digital era, inanimate objects have gained the capability to connect and engage with each other via the Internet. 
This phenomenon has given rise to a dynamic network of interconnected objects known as the \emph{Internet of Things} (IoT).
%
This revolution is further driven by the advent of telecommunication technologies such as 5G, offering remarkable bandwidth ranging from 100MB to 1GB per second and a mere 10 millisecond latency~\cite{ITU2015IMTVision}. 
This new larger bandwidth offers unprecedented opportunities for smart objects, especially those relying on computer vision for autonomous navigation, allowing real-time sharing of recorded images or videos through the network.

\begin{figure}
    \centering    \includegraphics[width=\columnwidth]{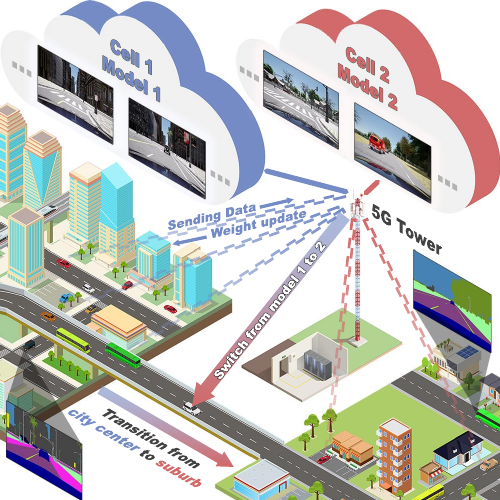}
    \caption{\textbf{Multi-Stream Cellular Test-time Adaptation (MSC-TTA) of real-time models.} We consider a set of agents (\eg, autonomous vehicles) evolving in a dynamic environment divided into cells (\eg, city center or suburb) that perform the same task (\eg, semantic segmentation) in real time on their own unlabeled data stream (\eg, recorded images) using an on-board model. We propose a first method in which agents share part of their data stream through an IoT network (\eg, a connection to a 5G tower). Cell-based lightweight models are then trained on the fly (in our case through an adaptive student-teacher method) and their weights are regularly broadcasted to the agents to improve their performance over time. When agents transitions between cells, the agent's model is immediately switched to the one of the new cell, effectively adapting the predictions of the transiting agent.}
    \label{fig:graphical}
\end{figure}

However, some critical computations need to be performed locally. 
For instance, autonomous vehicles should analyze their environment and take appropriate actions despite a loss of connection to the network.
This forces smart objects to include an on-board processing unit, especially for autonomous navigation. 
Due to limitations on battery capacities, these processing units are often limited in their computation capabilities.
Furthermore, the entire processing power can not just be dedicated to the analysis of the environment, but also needs to ensure all other critical functions (\eg, risk assessment, navigation system, \etc ).

Considering these limitations, the deployment of lightweight artificial intelligence models analyzing the environment becomes essential. 
Lightweight models offer the advantage of high inference speed, meeting the real-time constraint, and low power consumption.
However, their representational capacity is limited~\cite{Goodfellow2016Deep} compared to larger state-of-the-art model, failing in effectively handling a wide range of data distributions or generalizing to unseen environments at test time~\cite{Cioppa2019ARTHuS}.
In the case of moving objects such as autonomous vehicles, data distribution shifts generally occur as the vehicles transition between different areas. 
Even in the case of static objects such as surveillance cameras, the encountered data distribution may be dynamic, due, for example, to changes in weather conditions or object density and occlusions.
%
Fortunately, autonomous vehicles are able to precisely position themselves through a combination of navigation systems, telecommunications, and sensors (such as IMUs).
It is therefore possible to get prior knowledge on the encountered data distribution such as location (\eg, the region of the world, the city, or the neighbourhood) or local weather and traffic information.
Hence, the environment can be divided into a set of \emph{cells}, representing different dynamic data distributions.
Also, multiple objects evolving in similar environments (\eg, a fleet of autonomous vehicles) sense the environment and collect \emph{multiple streams}, allowing to sample the changing data distributions within each cell more quickly and comprehensively.

In this work, we propose a first \emph{Multi-Stream Cellular Test-Time Adaptation (MSC-TTA)} setup in which a fleet of connected objects, called \emph{agents}, adapt on the fly their model to their data stream with distribution shifts.
Then, we propose a first real-time method on our MSC-TTA setup based on an adaptive student-teacher online training strategy that leverages the division of the environment into different cells.
Specifically, each agent analyses its own data stream on board using a lightweight student model and offloads the heavy teacher inference process and student training remotely (\eg, on the cloud). 
As shown in Figure~\ref{fig:graphical}, data is collected by all agents and aggregated to train specialized student models for each cell. 
Finally, we study our new MSC-TTA setup in the practical real-world case of autonomous cars evolving in dynamic environments divided in different cells based on location (\eg, urban, suburbs, countryside, \etc) and weather conditions (\eg, sunny, rainy, foggy). To support our experiments and allow future benchmarking, we generate and publicly release a new large-scale synthetic semantic segmentation dataset based on the CARLA simulator~\cite{Dosovitskiy2017CARLA} called \emph{DADE}, and show improved performance of our proposed multi-stream and cell-based method over a single stream baseline.
We believe that this new multi-stream cellular test-time adaptation setup will open research possibilities for the combined use of computer vision and machine learning technologies in the IoT and 5G eras, as provisioned in 5G roadmaps~\cite{Linget2022AVsionary}. 

\mysection{Contributions.} We summarize our contributions as follows.
\textbf{(i)} We define a new Multi-Stream Cellular Test-Time Adaptation (MSC-TTA) setup in which models adapt on the fly to a dynamic environment divided into cells.
\textbf{(ii)} We propose a novel real-time adaptive student-teacher method to aggregate knowledge across different agents evolving in the same cell.
\textbf{(iii)} We generate and release a new synthetic dataset, called \emph{DADE}, for the semantic segmentation task on board autonomous vehicles and show improved performance of our proposed method compared to the baseline.

\section{Related Work}\label{sec:relatedwork}

\subsection{Online learning}
Online learning is a well-studied setup~\cite{Hazan2016Introduction, ShalevShwartz2011Online, Hoi2021Online, Orabona2019AModern-arxiv, CesaBianchi2021Online} defined as a game between a learner and an environment generating a stream of data. Based on past and current data generated by the environment, the learner tries to sequentially predict labels on the stream. At each step, the true label of the data is revealed and compared to the prediction of the learner. The learner then receives a regret score, used to penalize its mistakes.
The learner's objective is thus to minimize future penalties by using previously observed data and labels. The field of online learning can count on multiple benchmarking datasets such as firehose~\cite{Hu2022Drinking} for language modeling and CLOC~\cite{Cai2021Online} and CLEAR~\cite{DBLP:conf/nips/LinSPR21} for image classification of objects whose representations evolved over the span of $10$ years. In practice, online learning is relevant when the true label becomes available as time goes by, \eg, for the task of forecasting~\cite{Yang2019Online, Liu2016Online,Wang2014Online}.
In this work, we assess an upper bound of our MSC-TTA method by extending the setup to multi-stream cellular online learning.

\subsection{Test-time adaptation}
Similarly, Test-Time Adaptation (TTA) aims to adapt a model on a data stream. However, the environment does not reveal the true label of previously observed data. Several setups, characterized by the data distribution of the stream, have been studied, such as Fully TTA~\cite{DBLP:conf/iclr/WangSLOD21}, Continual TTA~\cite{Wang2022Continual},
Non-i.i.d TTA~\cite{Gong2022NOTE-arxiv}, or Practical TTA~\cite{Yuan2023Robust}, in which the data stream contains distribution changes and correlated samples. 
These setups are suited for real-world applications, where the true labels are unavailable at test time. However, previous works only consider a single stream of data. In this work, we go further, by proposing a setup for multiple streams and introducing prior knowledge on cross-stream data distribution through the division of the environment into cells. In addition, our methodology brings a real-time aspect, a feature often overlooked in previous setups.

To leverage the information in the data stream, multiple methods have been developed~\cite{Liang2023AComprehensive-arxiv}. Some works adapt the model's parameters by either fitting the batch normalization layers to the target domain \cite{Mirza2022TheNorm, Li2016Revisiting-arxiv, NEURIPS2020_85690f81}, training the model with auxiliary tasks \cite{DInnocente2020OneShot, DBLP:conf/icml/SunWLMEH20}, or fine-tuning it using unsupervised objectives \cite{DBLP:conf/nips/ZhangLF22, Reddy2022Master, DBLP:conf/iclr/WangSLOD21}. Some other works adapt the input data~\cite{Karani2021Testtime, He2020Self, 930081c8651747709037d63e19adc595, Gao2023Back} or weight the predictions of multiple models depending on the test distribution~\cite{Wang2021Efficient, Foll2022Gated-arxiv}. 
However, few works ensure that the adaptation is real time.

In fact, in real-world applications, the model needs to adapt within limited time to leverage all samples of the data stream, due to finite computing capabilities. Alfara~\etal~\cite{Alfarra2023Revisiting-arxiv} recently proposed an evaluation protocol to compare TTA methods under those constraints. 
To satisfy the real-time constraint, some works proposed a student-teacher architecture with a lightweight student model~\cite{Mullapudi2019Online, Cioppa2019ARTHuS}.
Specifically, \emph{ARTHuS}~\cite{Cioppa2019ARTHuS} proposed a first real-time method in which a lightweight student model is adapted on an unlabeled data stream at test time using pseudo-labels produced by a state-of-the-art but computation-expensive teacher model.
The real-time constraint of the system is ensured by asynchronously processing the student and teacher inference and training at different frame rates.
The fast lightweight student model therefore trains online on the changing data distributions using the teacher's slow predictions.
However, in the case of rapid domain shifts, the student needs several batches to adapt.
Houyon~\etal~\cite{Houyon2023Online} later tackled this issue by incorporating continual learning methods in the student online training to avoid catastrophic forgetting in the case of cyclic domain shifts.
Nevertheless, in the case of multiple objects (\eg, autonomous vehicles), each data stream is treated independently. In this work, we extend ARTHuS~\cite{Cioppa2019ARTHuS} to multiple data streams and cell-divided environments.

\subsection{Autonomous driving}
Autonomous Vehicles (AVs) rely on advanced sensor arrays, high-resolution cameras and on-board computing power to perceive the environment and make informed decisions to navigate safely. Nowadays, perception is largely based on artificial intelligence and involves several computer vision tasks such as semantic segmentation~\cite{Kirillov2020PointRend, Xie2021SegFormer, Yu2021BiSeNet, Zheng2021Rethinking}, object detection~\cite{Bochkovskiy2020YOLOv4-arxiv, Li2022Time3D} or depth estimation~\cite{Chang2018Pyramid, Godard2019Digging, Fonder2022Parallax}. However, the road to fully self-driving cars remains challenging.
For instance, it is still complex to operate AVs in diverse environments, such as varying weather conditions, traffic patterns and other unforeseen scenarios, and to process large amount of data while optimizing energy consumption in Electric Vehicles (EVs). 

To adapt to several environments, some methods use domain adaptation strategies~\cite{Wang2023Dynamically, Kennerley20232PCNet, Pierard2023Mixture} to enhance system versatility and reliability. 
Also, cloud computing~\cite{Schafhalter2023Leveraging-arxiv} or multi-access edge computing (MEC)~\cite{Liu2019Edge, Yang2021Edge} provide the computational power and storage capacity for real-time data processing, enhancing energy efficiency and improving EV mileage.
Similarly to MEC, our proposed method employs a hybrid approach. On-board processing handles immediate, low-latency operations, while resource-intensive computations are offloaded to external servers. 
Specifically, the heavy offloaded computations rely on the multiple streams of the fleet, while on-board, lightweight real-time perception is performed using models trained in the cloud, guaranteeing adaptability in dynamic environments.

\section{Methodology}\label{sec:method}


\begin{figure*}
    \centering
    \includegraphics[width=\linewidth]{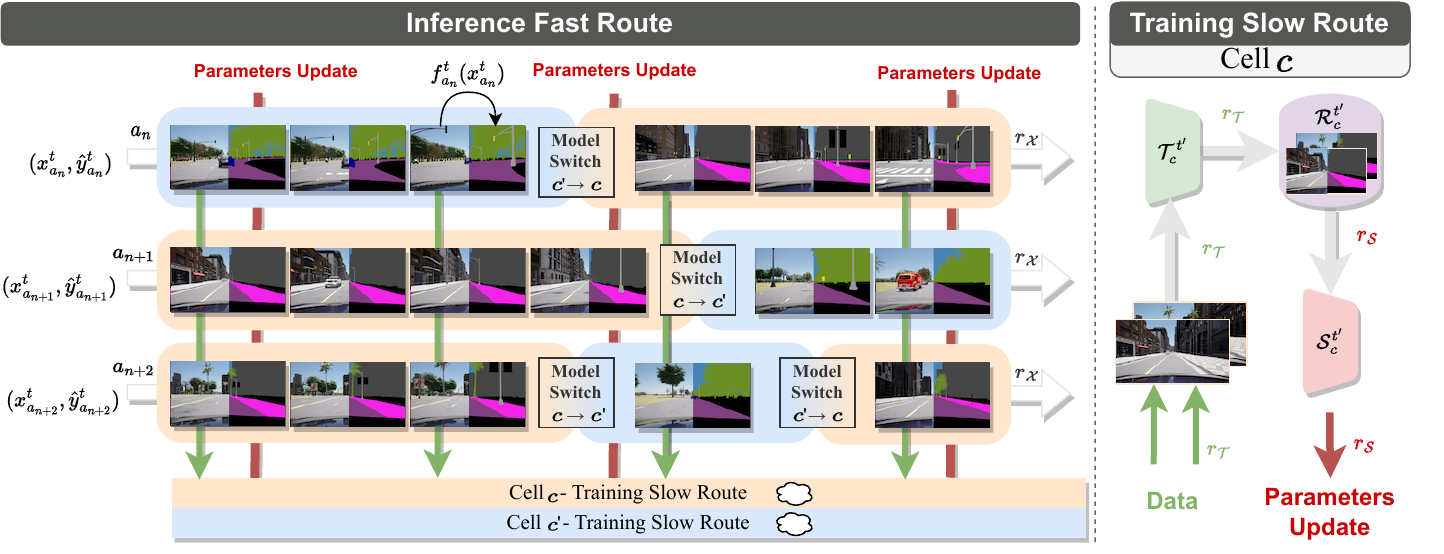}
    \caption{\textbf{Pipeline of our multi-stream cellular test-time adaptation of real-time models.} 
    Our method is composed of a fast route for inference and a slow route for online training, as defined in~\cite{Cioppa2019ARTHuS,Houyon2023Online}. 
    In the fast route, each agent $\agent{\agentIndex}$ processes a stream of data samples $\sample{\agent{\agentIndex}}{\time}$ and predicts labels $\prediction{\agent{\agentIndex}}{\time}=\model{\agent{\agentIndex}}{\time}(\sample{\agent{\agentIndex}}{\time})$ in real time (\ie, at the data stream rate $\framerate{\datastream{}{}}$).
    Agents located within a cell $\environmentAlone{\environmentIndex}{}$ send a subset of their data samples at a slower rate $\framerate{\teacher{}{}})$ to a slow route operating on a remote server (\eg, on the cloud) dedicated for each cell. 
    In the slow route, a teacher model $\teacher{\environmentAlone{\environmentIndex}{}}{\time'}$ predicts pseudo labels on the received data and stores them in a replay buffer $\replayBuffer{\environmentAlone{\environmentIndex}{}}{\time'}$. 
    The replay buffer is then used to train on the fly a cell-specific student model $\student{\environmentAlone{\environmentIndex}{}}{\time'}$ at a rate $\framerate{\student{}{}}$. 
    After each training epoch on the replay buffer, the parameters of $\student{\environmentAlone{\environmentIndex}{}}{}$ are transferred to all agent models $\model{\agent{\agentIndex}}{}$ located within that cell. 
    Finally, agents transiting between two cells have their model switched instantly.}
    \label{fig:pipeline}
\end{figure*}

\subsection{Problem statement}

Given a finite set of $\agentSize$ agents $\agent{\agentIndex}$ forming a connected fleet $\allAgents$, our proposed Multi-Stream Cellular Test-Time Adaptation (MSC-TTA) aims to adapt over time $\time \in \{0,...,T\}$ each agent's model $\model{\agent{\agentIndex}}{\time}\in\allModels$, pretrained on any source domain to perform a task $\task$, to the agent's data stream $\datastream{\agent{\agentIndex}}{ }\in\allDatastreams$ of online unlabeled samples $\datastream{\agent{\agentIndex}}{ }=\sample{\agent{\agentIndex}}{0}, \sample{\agent{\agentIndex}}{1},...,\sample{\agent{\agentIndex}}{\time},...,\sample{\agent{\agentIndex}}{\lastTime}$.
As in \cite{Yuan2023Robust}, the samples are drawn from a distribution $\distribution{\agent{\agentIndex}}{}\in\allDistributions$ shifting over time following $\distribution{\agent{\agentIndex}}{0},\distribution{\agent{\agentIndex}}{1},...,\distribution{\agent{\agentIndex}}{\time},...,\distribution{\agent{\agentIndex}}{\lastTime}$, in which consecutive samples $\sample{\agent{\agentIndex}}{\time-1},\sample{\agent{\agentIndex}}{\time}, \sample{\agent{\agentIndex}}{\time+1}$ may be highly correlated.
At time $\time$, the model $\model{\agent{\agentIndex}}{t}$ receives a batch of unlabeled samples $\batch{\agent{\agentIndex}}{\time} = \sample{\agent{\agentIndex}}{\time}, \sample{\agent{\agentIndex}}{\time+1},..., \sample{\agent{\agentIndex}}{\time+(\batchsize-1)}$, where $\batchsize$ is the batch size, on which it makes predictions. Each model $\model{\agent{\agentIndex}}{\time}$ may be adapted to the current batch $\batch{\agent{\agentIndex}}{\time}$ by accumulating knowledge from previous samples of the multiple streams forming the following hyperspace $\cup\sample{\agent{\agentIndex}}{\time'}, \forall {\agent{\agentIndex}\in\allAgents \times{\time'<(\time+\batchsize)}}$.
%
Let us note that samples $\sample{\agent{\agentIndex}}{\time}$ may be unavailable for some time $\time$ for some agent $\agent{\agentIndex}$. 
This setup describes the general case of multiple sensors recording data streams and performing the same task, \eg, surveillance cameras placed in one or several cities on which crowd counting or car segmentation needs to be performed, with no assumptions on where the cameras are placed. 

To include cross-stream prior knowledge on data distributions, we consider the general case in which the agents evolve inside a dynamic environment split into a non-overlapping set of $\environmentSize$ cells $\environmentAlone{\environmentIndex}{}\in\allEnvironments$.
We suppose that, at time $\time$, each agent is located within one cell such that $\environment{\agent{\agentIndex}}{\time}=\environmentAlone{\environmentIndex}{}\in\allEnvironments$, with agents being able to transition between cells over time. 
The cells $\environmentAlone{\environmentIndex}{ }$ are predefined by a set of rules (\eg, based on the location, the weather, etc.) such that the expected data distribution of agents evolving in the same cell is similar, \ie, 
$\distribution{\agent{\agentIndex}}{\time} \approx \distribution{\agent{\agentIndexb}}{\time}$ if $\environment{\agent{\agentIndex}}{\time} = \environment{\agent{\agentIndexb}}{\time}$.
Our setup therefore allows the different data streams to share common data distribution properties at times that can be leveraged to effectively adapt the models.
Naturally, in practice, this assumption may fail if the cells are incorrectly defined or estimated.
This setup is particularly interesting in the real-world case of autonomous driving, in which vehicles evolve in different locations (\eg, city centers, suburbs, highway, \etc) that they analyze through various sensors. Also, vehicles driving in the same environment may leverage the multiple streams of the fleet to better assess and adapt to the environment.
%
%

Let us note that considering the special case of $\agentSize=1$ and $\environmentSize=1$ falls back to the original PTTA setup of Yuan~\etal~\cite{Yuan2023Robust} in which a single model is adapted to its data stream. The case of $\agentSize\geq1$ and $\environmentSize=1$ represents a Multi-Stream Test-Time Adaptation (MS-TTA) setup without division of the environment. Finally, the case of $\environmentSize=1$ and $\model{\agent{\agentIndex}}{\time} = \model{ }{\time}$, \ie, in which a single model is adapted for all streams without prior knowledge on the environments, corresponds to a TTA setup in which samples from multiple streams are combined in the batch.
In the following, we describe our adaptive method for the general case $\agentSize \geq 1$ and $\environmentSize \geq 1$. To stay close to a real-world scenario, we add an extra real-time constraint on the method, \ie, no delay accumulation or sample skipping when processing the multiple data streams.

\begin{table*}
    \centering
    \caption{\textbf{Mean IoU performance on DADE-static weather.} The MSC-OL setup leverages the CARLA segmentation masks as pseudo labels while the MSC-TTA setup leverages pseudo labels from the teacher model. We compare several pretraining, scenarios, and adaptive (\checkmark) versus frozen (\snow) models. For each pretraining, the best score is shown in \textbf{bold} and the second is \uline{underlined}.}
    \resizebox{\textwidth}{!}{
    \begin{tabular}{c|c|c||c|c|c|c||c|c|c|c}
        \multicolumn{3}{c||}{} &  \multicolumn{4}{c||}{\textbf{Multi-stream cellular online learning}} &  \multicolumn{4}{c}{\textbf{Multi-stream cellular test-time adaptation}} \\ \cmidrule(lr){4-11}
        \multicolumn{3}{c||}{} &  \multicolumn{2}{c|}{\textbf{mIoU imminent}} &  \multicolumn{2}{c||}{\textbf{mIoU future}} & \multicolumn{2}{c|}{\textbf{mIoU imminent}} &  \multicolumn{2}{c}{\textbf{mIoU future}}\\ \midrule
        \textbf{Pretraining}     & \textbf{Scenario}           & \textbf{Adapt}                 & 3 hours & Last hour & 3 hours & Last hour & 3 hours & Last hour & 3 hours & Last hour\\ \midrule
        \multirow{2}{*}{Cityscapes~\cite{Cordts2016The}}         & Student           & \snow                    & .214 & .218 & .214 & .218 & .214 & .218 & .214 & .218\\
                                                 & Teacher           & \snow                    & .668 & .671 & .668 & .671 & .668 & .671 & .668 & .671\\ \midrule
        \multirow{4}{*}{\textit{Scratch}}         & \textit{Baseline}~\cite{Cioppa2019ARTHuS}           & \checkmark                    & .223 & .249 & .208 & .231 & .274 & .309 & .244 & .285\\
                & \textit{Baseline}+MIR~\cite{Houyon2023Online}           & \checkmark                    & .173 & .194 & .164 & .188 & .181 & .201 & .171 & .195\\
                 & \textit{Common}            & \checkmark                    & \uline{.338} & \uline{.483} & \uline{.316} & \uline{.461} & \uline{.340} & \uline{.363} & \uline{.327} & \uline{.373}\\
                 & \textit{Spatial}            & \checkmark                    & \textbf{.353} & \textbf{.513} & \textbf{.328} & \textbf{.485} & \textbf{.368} & \textbf{.440} & \textbf{.351} & \textbf{.413}\\ \midrule
        \multirow{5}{*}{\textit{General}}         & \textit{Baseline}~\cite{Cioppa2019ARTHuS}           & \checkmark                    & .435 & .442 & .415 & .446 & .422 & .442 & .397 & .425\\
        & \textit{Baseline}+MIR~\cite{Houyon2023Online}           & \checkmark                    & .650 & .656 & .614 & .626 & .417 & .432 & .401 & .423\\
                 & \textit{Common}            & \checkmark                    & \textbf{.702} & \uline{.696} & \textbf{.673} & \uline{.692} & \textbf{.474} & \textbf{.517} & \uline{.461} & \uline{.501}\\           
                 & \textit{Spatial}            & \checkmark                    & \uline{.700} & \textbf{.701} & \uline{.660} & \textbf{.701} & \uline{.470} & \textbf{.517} & \textbf{.462} & \textbf{.505}\\
                 & \textit{Common}            & \snow                             & .650 & .658 & .650 & .658 & .454 & \uline{.450} & .454 & .450\\\midrule
        \multirow{2}{*}{\textit{Cell}}            & \textit{Spatial}            & \checkmark                    & \textbf{.658} & \textbf{.681} & \uline{.597} & \textbf{.682} & \textbf{.552} & \uline{.567} & \uline{.522} & \uline{.556}\\      
                 & \textit{Spatial}            & \snow                             & \uline{.634} & \uline{.660} & \textbf{.634} & \uline{.660} & \uline{.544} & \textbf{.572} & \textbf{.544} & \textbf{.572}\\
        \bottomrule

    \end{tabular}}
    
    \label{tab:agg_sw}
\end{table*}
\begin{figure*}
    \centering
    \includegraphics[width=\linewidth]{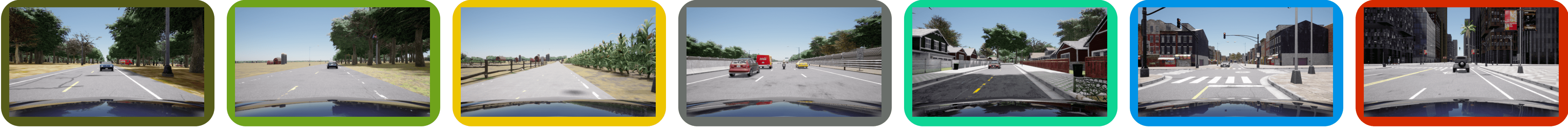} 
    \caption{\textbf{Images of the different locations in our dataset.} We define $7$ different locations that are defined based on the GNSS data. From left to right: \textcolor{forest}{forest}, \textcolor{countryside}{countryside}, \textcolor{rural_farmland}{rural farmland}, \textcolor{highway}{highway}, \textcolor{low_density_residential}{low density residential}, \textcolor{community_buildings}{community buildings}, and \textcolor{high_density_residential}{high density residential}.}
    \label{fig:dataset-zones}
\end{figure*}
\subsection{Multi-stream cellular test-time adaptation}

Our method, illustrated in Figure~\ref{fig:pipeline}, produces a stream of predictions for every agent following $\prediction{\agent{\agentIndex}}{\time}=\model{\agent{\agentIndex}}{\time}(\sample{\agent{\agentIndex}}{\time})$, with the model $\model{\agent{\agentIndex}}{\time}$ operating in real time (\ie, at the rate $\framerate{\datastream{}{}}$) on the data stream $\datastream{\agent{\agentIndex}}{ }$. 
To do so, we extend the adaptive real-time student-teacher method, ARTHuS,  of Cioppa~\etal~\cite{Cioppa2019ARTHuS}, in which a lightweight student model $\student{ }{ }$ is adapted on-the-fly using pseudo labels produced by a state-of-the-art but computation-expensive teacher model $\teacher{ }{ }$.
Particularly, we leverage the multiple streams and the division of the environment into cells.
We allow agents evolving within the same cell to share their own data stream to produce a cell-specific data stream $\datastream{\environmentAlone{\environmentIndex}{ }}{\time'} = \cup \sample{\agent{\agentIndex}}{\time'}, \forall \agent{\agentIndex} \mid \environment{\agent{\agentIndex}}{\time'} = \environmentAlone{\environmentIndex}{}$ at a frame rate $\framerate{\teacher{ }{ }}$, producing samples $\sample{\environmentAlone{\environmentIndex}{ }}{\time'}$.

Our method is composed of a fast route and a slow route.
In the fast route (inference), student models for each agent produce predictions  $\prediction{\agent{\agentIndex}}{\time} = \student{\agent{\agentIndex}}{\time}(\sample{\agent{\agentIndex}}{\time}) = \model{\agent{\agentIndex}}{\time} (\sample{\agent{\agentIndex}}{\time})$ at the rate $\framerate{\datastream{}{}}$. In parallel in the slow route (training), a slow but high-performance teacher model $\teacher{\environmentAlone{\environmentIndex}{}}{\time'}$ for each cell produces pseudo-ground truths $\pseudo{\environmentAlone{\environmentIndex}{}}{\time'} = \teacher{\environmentAlone{\environmentIndex}{}}{\time'}(\sample{\environmentAlone{\environmentIndex}{ }}{\time'})$ at the rate $\framerate{\teacher{ }{ }}$ on the cell data streams. The pair of data $(\sample{\environmentAlone{\environmentIndex}{ }}{\time'},\pseudo{\environmentAlone{\environmentIndex}{}}{\time'})$ are then stored in a replay buffer $\replayBuffer{\environmentAlone{\environmentIndex}{}}{\time'}$ of size $\replayBufferSize$ using a First-In-First-Out (FIFO) strategy. One student network per cell $\student{\environmentAlone{\environmentIndex}{}}{\time'}$ is trained on the updated replay buffer $\replayBuffer{\environmentAlone{\environmentIndex}{}}{\time'}$ using a loss function
\begin{equation}
\loss = \sum_{\replayIndex=1}^{\replayBufferSize}\distance(\student{\environmentAlone{\environmentIndex}{}}{\time'}(\sample{\environmentAlone{\environmentIndex}{}}{\replayIndex}),\pseudo{\environmentAlone{\environmentIndex}{}}{\replayIndex}),
\end{equation} where $\distance$ is a dissimilarity measure suited for task $\task$.
After training for one epoch on the replay buffer, the weights of students in the fast route are updated with the weight of the environment students in the slow route such that $\student{\agent{\agentIndex}}{\time} = \student{\environmentAlone{\environmentIndex}{}}{\time}, \forall \agent{\agentIndex} | \environment{\agent{\agentIndex}}{\time} = \environmentAlone{\environmentIndex}{}$, at a slower rate $\framerate{\student{}{}}$. Since the slow route gathers information from several agents, the heavy teacher inference and student training processes can be offloaded to a dedicated server (\eg, on the cloud). 
Hence, agents only perform the real-time inference with a lightweight model, greatly reducing computation requirements and saving precious battery power in the case of autonomous vehicles.
Finally, considering the special case $\environmentSize=\agentSize$ with each agent defining its own cell is equivalent to the original ARTHuS method~\cite{Cioppa2019ARTHuS}, serving as baseline in our experiments.

\section{Experiments}\label{sec:experiments}

\subsection{Dataset}

To support our experiments, we generate and release the Driving Agents in Dynamic Environments (DADE) dataset, based on the CARLA simulator. DADE is tailored for the online training and evaluation of semantic segmentation methods in the context of autonomous driving agents navigating dynamic environments The first part of DADE contains $100$ video sequences of agents evolving in $7$ connected locations illustrated in Figure~\ref{fig:dataset-zones}, with static weather conditions (clear day). 
The second part contains $300$ video sequences in the same locations with dynamic weather conditions (clear, rainy, and foggy), during day and night. 
We provide video sequences, semantic segmentation masks, Global Navigation Satellite System (GNSS) data, and weather information. Each sequence is acquired by an agent within a $5$-hours time frame. 
The first two hours are used for pretraining and the remaining three for adaptation.


To the best of our knowledge, our dataset, large of 150 GBytes, is the first to provide long videos of multiple agents evolving in diverse driving locations and weather conditions with ground truth labels for the task of semantic segmentation. 
Our video sequences contain between $188$ and $7{,}200$ frames acquired at $1$ frame per second (fps), with an average sequence length of $40$ minutes.  Existing datasets, such as~\cite{Cordts2016The, Sun2022SHIFT, Maddern20171Year}, feature short video sequences, lack multi-agent perspectives, do not include ground truth data, or lack a diverse range of weather conditions.
More information about our DADE dataset may be found in the appendix.

\subsection{Experimental settings}

\begin{table*}
    \centering
    \caption{\textbf{Mean IoU performance on DADE-dynamic weather.} The MSC-OL setup leverages the CARLA segmentation masks as pseudo labels while the MSC-TTA setup leverages pseudo labels from the teacher model. We compare several pretraining, scenarios, and adaptive (\checkmark) versus frozen (\snow) models. For each pretraining, the best score is shown in \textbf{bold} and the second is \uline{underlined}.}
    \resizebox{\textwidth}{!}{
    
    \begin{tabular}{c|c|c||c|c|c|c||c|c|c|c}
        \multicolumn{3}{c||}{} &  \multicolumn{4}{c||}{\textbf{Multi-stream cellular online learning}} &  \multicolumn{4}{c}{\textbf{Multi-stream cellular test-time adaptation}} \\ \cmidrule(lr){4-11}
        \multicolumn{3}{c||}{} &  \multicolumn{2}{c|}{\textbf{mIoU imminent}} &  \multicolumn{2}{c||}{\textbf{mIoU future}} & \multicolumn{2}{c|}{\textbf{mIoU imminent}} &  \multicolumn{2}{c}{\textbf{mIoU future}}\\ \midrule
        \textbf{Pretraining}     & \textbf{Scenario}           & \textbf{Adapt}                 & 3 hours & Last hour & 3 hours & Last hour & 3 hours & Last hour & 3 hours & Last hour\\ \midrule
        \multirow{2}{*}{Cityscapes~\cite{Cordts2016The}}         & Student           & \snow                    & .159 & .130 & .159 & .130 & .159 & .130 & .159 & .130\\
                                                 & Teacher           & \snow                    & .611 & .542 & .611 & .542 & .611 & .542 & .611 & .542\\ \midrule
        \multirow{7}{*}{\textit{Scratch}}         & \textit{Baseline}~\cite{Cioppa2019ARTHuS}           & \checkmark                    & .204 & .197 & .167 & .167 & .212 & .190 & .173 & .173\\
        & \textit{Baseline}+MIR~\cite{Houyon2023Online}           & \checkmark                    & .144 & .137 & .125 & .118 & .147 & .133 & .129 & .110\\
                 & \textit{Common}            & \checkmark                    & \uline{.278} & \uline{.352} & \uline{.249} & \uline{.323} & \uline{.278} & \uline{.257} & \uline{.253} & \uline{.243}\\
                 & \textit{Spatial}            & \checkmark                    & \textbf{.307} & \textbf{.397} & \textbf{.269} & \textbf{.358} & \textbf{.312} & \textbf{.300} & \textbf{.278} & \textbf{.276}\\
                 & \textit{Weather}            & \checkmark                    & .226 & .295 & .199 & .279 & .227 & .216 & .202 & .197\\
                 & \textit{Daylight}           & \checkmark                    & .245 & .279 & .176 & .259 & .182 & .198 & .150 & .184\\
                 & \textit{Specific}           & \checkmark                    & .22  & .204 & .203 & .187 & .233 & .186 & .218 & .166\\ \midrule
        \multirow{8}{*}{\textit{General}}         & \textit{Baseline}~\cite{Cioppa2019ARTHuS}           & \checkmark                   & .581 & .546 & .502 & .502 & .471 & .406 & .409 & \uline{.409}\\
                 & \textit{Baseline}+MIR~\cite{Houyon2023Online}           & \checkmark                    & .567 & .531 & .527 & .480 & .455 & .386 & .427 & .347\\
                 & \textit{Common}            & \checkmark                    & \uline{.644} & .595 & \uline{.613} & .565 & .506 & .427 & \uline{.483} & .405\\
                 & \textit{Spatial}            & \checkmark                    & \textbf{.654} & \textbf{.622} & .606 & \textbf{.589} & \textbf{.516} & \textbf{.442} & .473 & .405\\
                 & \textit{Weather}            & \checkmark                    & .641 & .586 & .611 & .562 & \uline{.507} & .429 & \textbf{.484} & .408\\
                 & \textit{Daylight}           & \checkmark                    & .636 & \uline{.603} & .572 & \uline{.585} & .498 & .430 & .477 & \textbf{.413}\\
                 & \textit{Specific}           & \checkmark                    & .632 & .602 & .596 & .559 & .500 & \uline{.437} & .471 & .393\\ 
                 & \textit{Common}            & \snow                             & .618 & .581 & \textbf{.618} & .581 & .476 & .403 & .476 & .403\\ \midrule
        \multirow{8}{*}{\textit{Cell}}            & \textit{Spatial}            & \checkmark                    & \textbf{.662} & \textbf{.642} & .609 & \uline{.590} & \textbf{.527} & \textbf{.461} & .484 & \textbf{.423}\\
                    & \textit{Weather}            & \checkmark                    & .634 & .580 & .607 & .551 & \uline{.509} & .427 & .483 & .409\\
                    & \textit{Daylight}           & \checkmark                    & \uline{.645} & .592 & \uline{.620} & .577 & .507 & .432 & \textbf{.488} & \uline{.415}\\
                    & \textit{Specific}           & \checkmark                    & .612 & .582 & .589 & .554 & .500 & \uline{.438} & \uline{.485} & .412\\
                    & \textit{Spatial}            & \snow                             & .642 & \uline{.606} & \textbf{.642} & \textbf{.606} & .488 & .409 & \textbf{.488} & .409\\
                    & \textit{Weather}            & \snow                             & .565 & .528 & .565 & .528 & .443 & .384 & .443 & .384\\
                    & \textit{Daylight}           & \snow                             & .563 & .485 & .563 & .485 & .421 & .362 & .421 & .362\\
                    & \textit{Specific}           & \snow                             & .447 & .400 & .447 & .400 & .349 & .298 & .349 & .298\\ \bottomrule
    \end{tabular}}
    
    \label{tab:agg_dw}
\end{table*}

\subsubsection{Environment division}
We consider six scenarios based on the division of the environment into cells. (1)~The \textit{Baseline} scenarios correspond to multiple independent streams on which independent agents adapt ($\environmentSize=\agentSize$), \ie, ARTHuS~\cite{Cioppa2019ARTHuS} and Houyon~\etal~\cite{Houyon2023Online}. (2)~The \textit{Common} scenario aggregates the multiple data streams into a single stream, on which one common model adapts ($\environmentSize=1$). (3)~The \textit{Spatial} scenario leverages the different locations of our dataset to split the environment into cells ($\environmentSize = 7$). (4)~The \textit{Weather} and (5)~\textit{Daylight} scenarios temporally divide the environment based on the weather ($\environmentSize = 3$: clear, rainy, foggy) and the time period ($\environmentSize = 2$: day, night). (6)~The \textit{Specific} scenario considers each combination of location, weather condition, and time period ($\environmentSize = 42$). 

\subsubsection{Pretraining}
We choose the same model architecture for all agent models $\model{\agent{\agentIndex}}{}$ and cell-specific student models $\student{\environmentAlone{\environmentIndex}{}}{}$. 
Following the work of Cioppa~\etal\cite{Cioppa2019ARTHuS}, we select TinyNet: a lightweight semantic segmentation model operating in real time. 
The pretraining set is divided into a training set and a validation set using a $90$-$10\%$ split.
For each scenario, we evaluate $3$ pretraining techniques. The \textit{General} pretraining consists in training the student model on all samples of the training set, irrespective of the division into cells. The \textit{Cell} pretraining considers a separate model for each cell $\environmentAlone{\environmentIndex}{ }$, trained on cell-specific samples. Finally, \textit{Scratch} assigns random weights (\ie, no pretraining). 
The models are pretrained with a learning rate of $10^{-4}$ using the Adam optimizer and the best performing model over the validation set is selected. The number of epochs is set to $3$ for the \textit{General} pretraining and scaled for each cell for the \textit{Cell} pretraining to match the number of backward passes and ensure a fair comparison.

\begin{figure*}
  \centering
  \begin{minipage}[b]{0.4\textwidth}
    \centering
    \includegraphics[width=\linewidth]{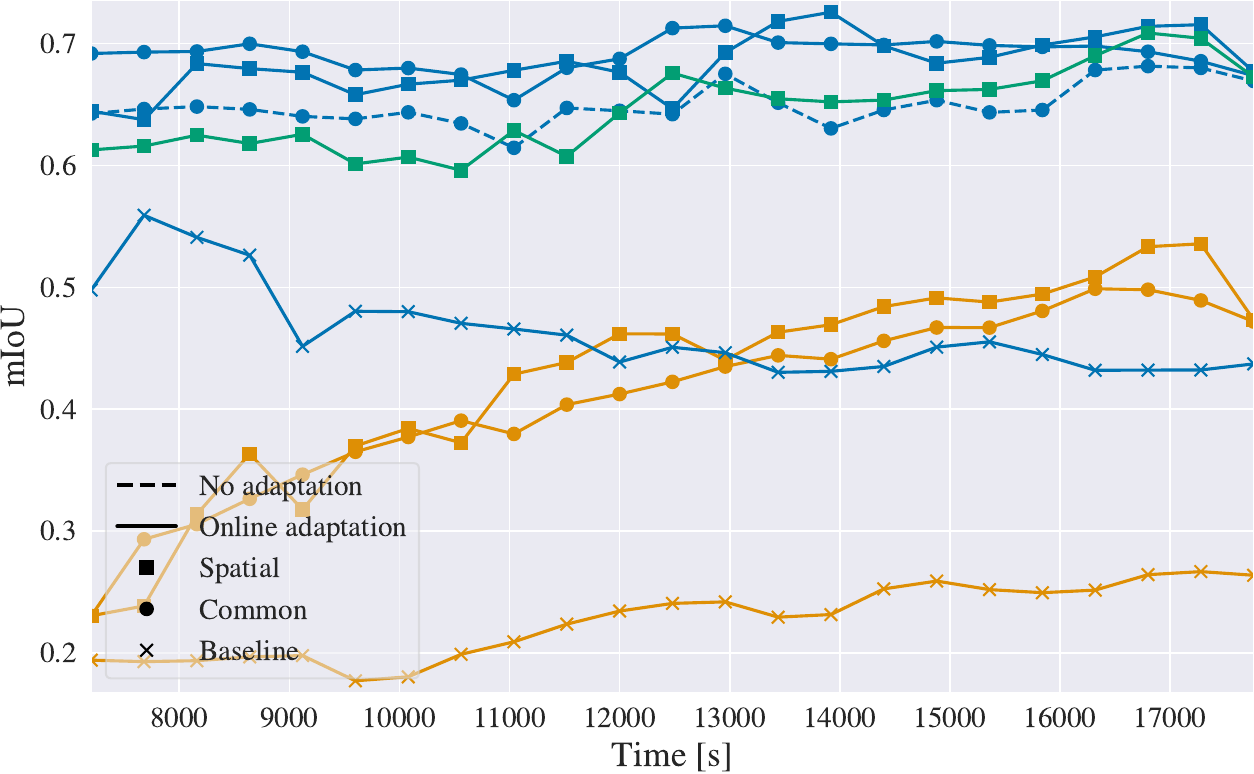}
  \end{minipage}
  \hspace{1.5cm}
  \begin{minipage}[b]{0.4\textwidth}
    \centering
    \includegraphics[width=\linewidth]{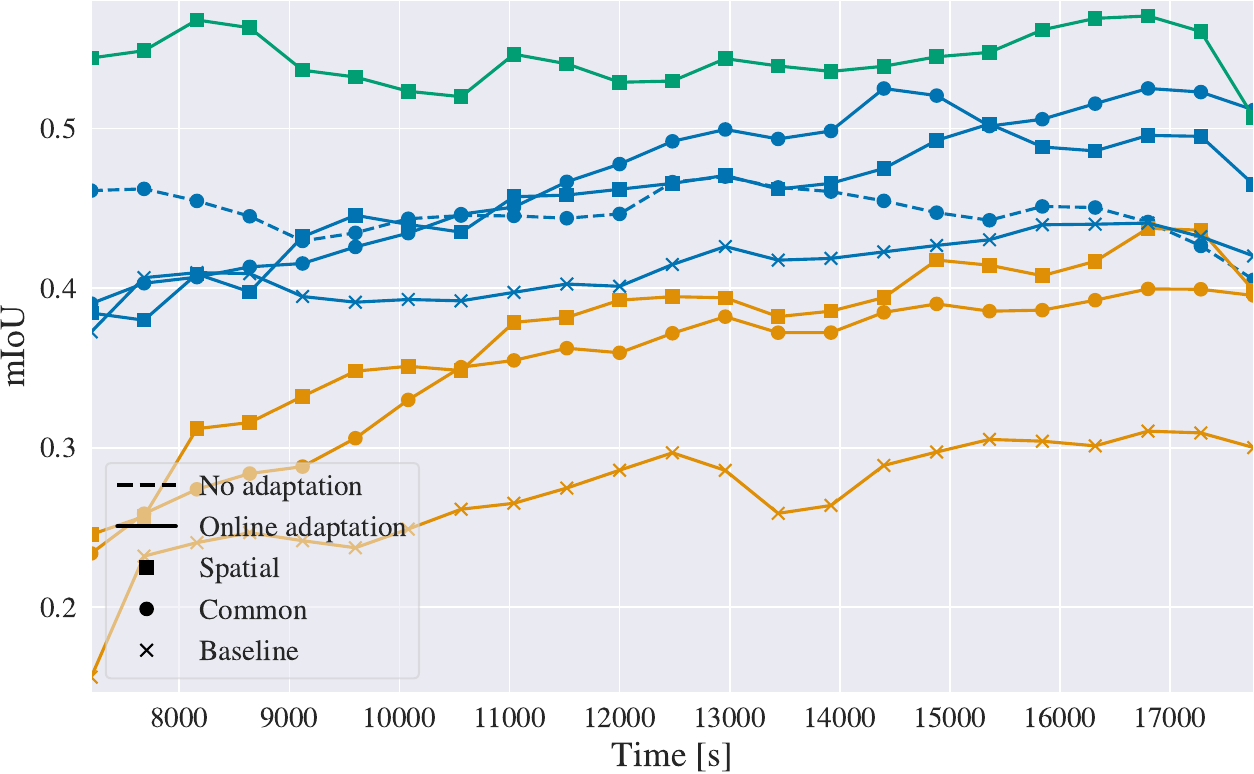}
    
  \end{minipage}
  \centering
  \begin{minipage}[b]{0.4\textwidth}
    \centering
    \includegraphics[width=\linewidth]{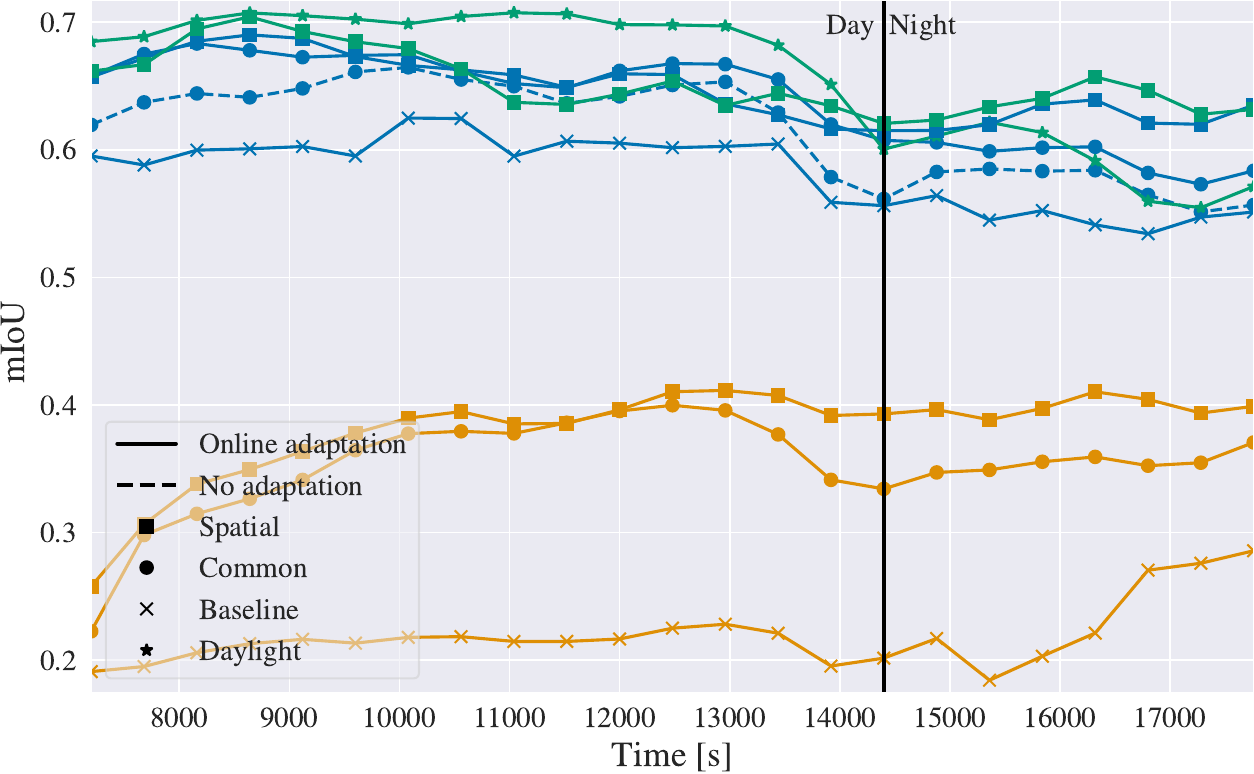}
  \end{minipage}
  \hspace{1.5cm}
  \begin{minipage}[b]{0.4\textwidth}
    \centering
    \includegraphics[width=\linewidth]{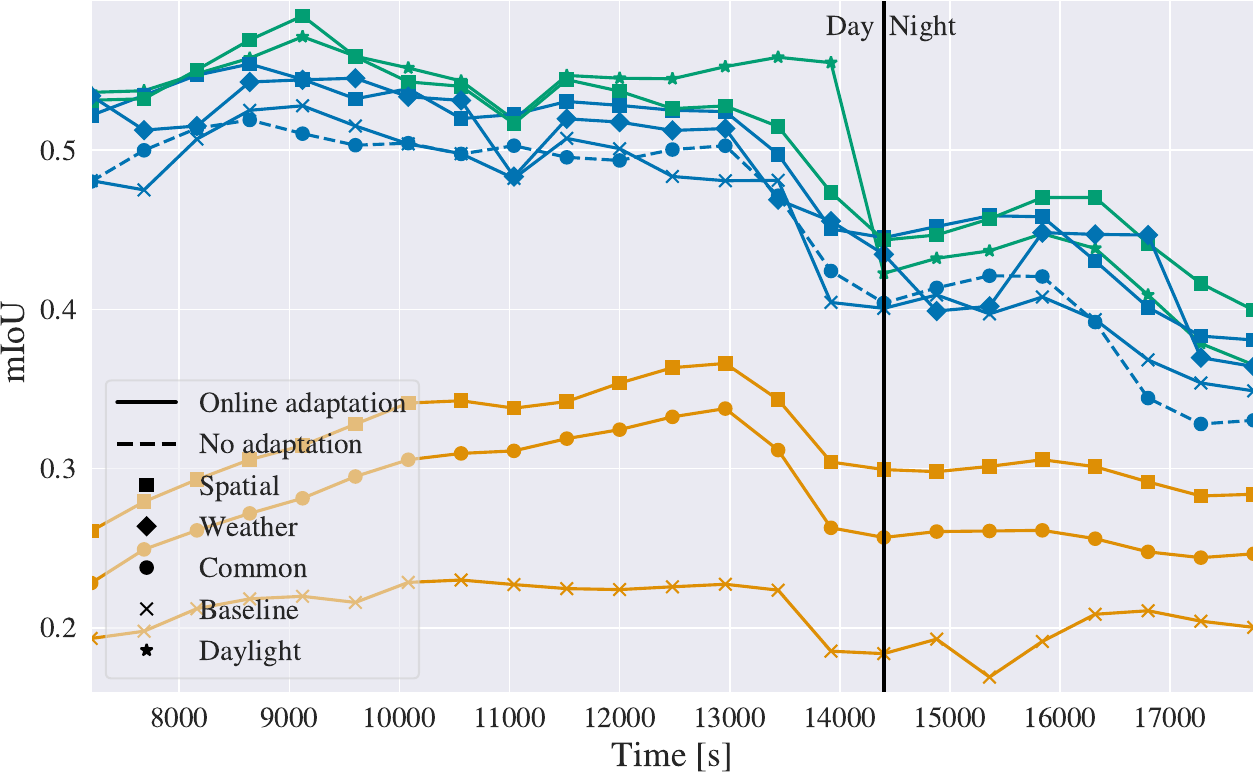}
    
  \end{minipage}
  \caption{\textbf{Evolution of the fleet performance over time on DADE-static weather (top) and DADE-dynamic weather (bottom).} Comparison of the performance in the MSC-OL setup (left) and the MSC-TTA (right) setup of the best adaptive settings along with the baseline for each pretraining (\textcolor{Scratch}{\textit{Scratch}}, \textcolor{General}{\textit{General}}, and \textcolor{Cell}{\textit{Cell}}). 
  } 
    \label{fig:exp_sw} 
\end{figure*}

\subsubsection{Testing}
For a given scenario and pretraining procedure, we compare the online performance (\ie, our adaptive method) with the offline performance (\ie, a frozen pretrained model). 
We choose the teacher model as a frozen state-of-the-art SegFormer~\cite{Xie2021SegFormer} model trained on Cityscapes~\cite{Cordts2016The} that produces pseudo labels at a rate of $\framerate{\teacher{ }{ }} = 1/3\,[\text{Hz}]$.
The replay buffers are chosen as FIFO buffers with a size $\replayBufferSize = 100$, updated at the same rate $\framerate{\teacher{ }{ }}$.
Finally, the cell-specific student models are trained online at a rate $\framerate{\student{}{}} = 1/30\,[\text{Hz}]$, with a learning rate of $10^{-4}$, batch size of $25$ with the Adam optimizer, and the cross-entropy loss. The model is only trained if the buffer contains new samples to prevent overfitting.

For the online evaluation, we aggregate the confusion matrices over a sliding window of $30\,[\text{s}]$ (imminent performance) for every agent in every cell and compute the mean-Intersection-over-Union (mIoU) as defined in Houyon~\etal~\cite{Houyon2023Online}. Additionally, we propose to evaluate the current model $5$ minutes in the future (future performance) to assess the capacity of the model to generalize to future samples.
Finally, we also compute the overall mIoU for the entire test set (3 hours) and for the last hour to assess the long-term performance.
As an upper bound, we also evaluate our method in a Multi-Stream Cellular Online Learning (MSC-OL) setup by replacing the teacher pseudo labels by the true ground-truth labels. We also compare our approach to the best method proposed in Houyon~\etal~\cite{Houyon2023Online}, which is equivalent to our \textit{Baseline} with a Maximal Interfered Retrieval (MIR) buffer, and report the offline performances of the frozen teacher and student models both trained on Cityscapes~\cite{Cordts2016The}. 
More information on the evaluation, such as details about the object classes, is provided in the appendix.

\subsection{Results}
\begin{figure*}
    \centering
    \begin{subfigure}{.17\linewidth}
        \centering
        \includegraphics[width=\linewidth]{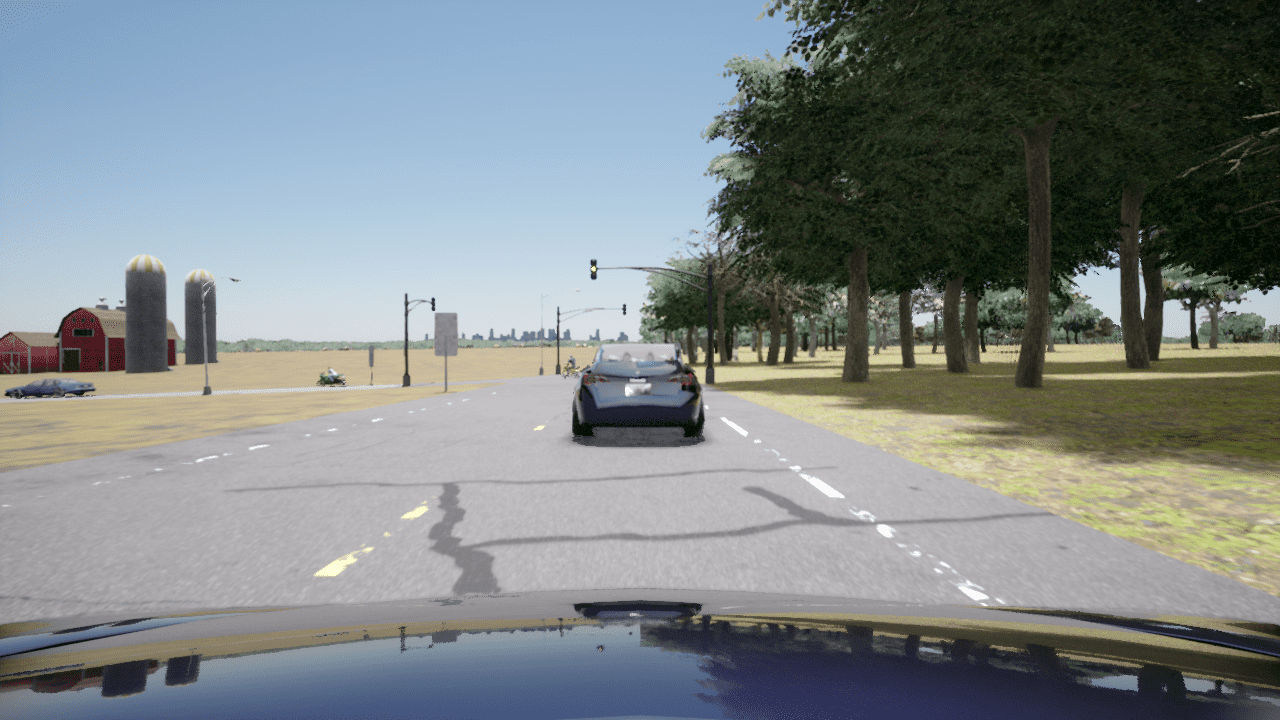}
    \end{subfigure}
    \begin{subfigure}{.17\linewidth}
        \centering
        \includegraphics[width=\linewidth]{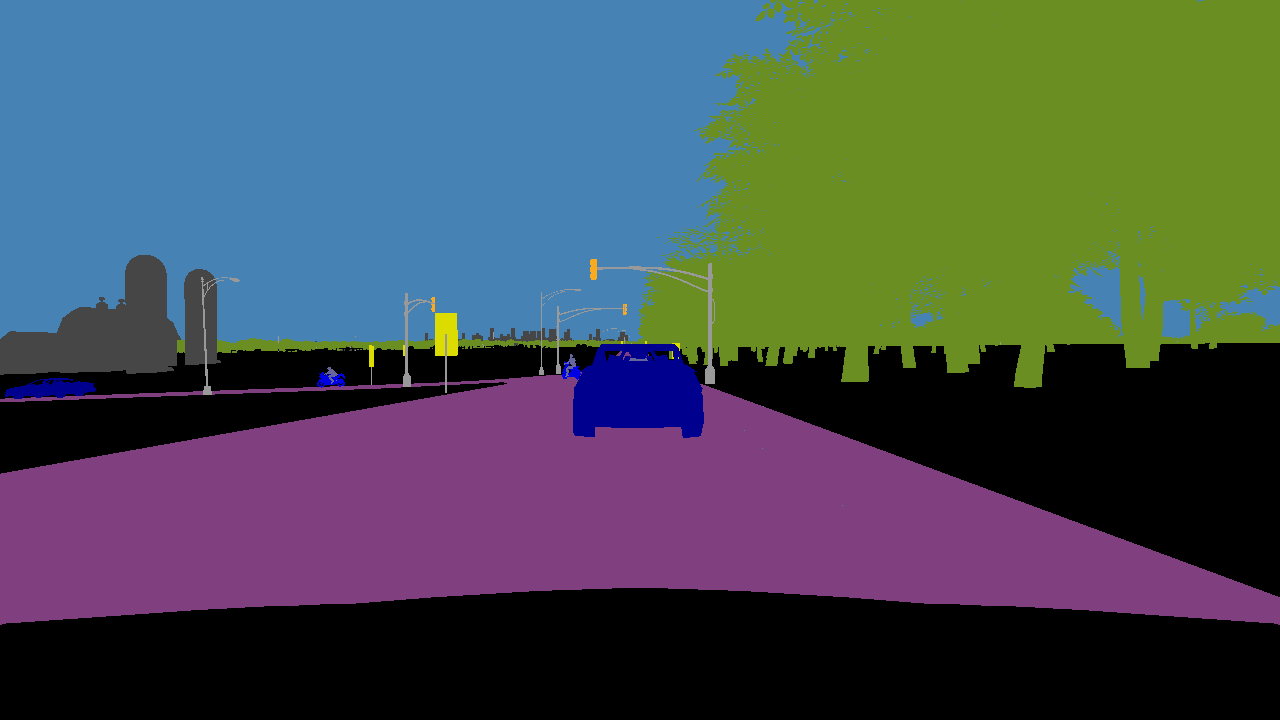}
    \end{subfigure}
    \begin{subfigure}{.17\linewidth}
        \centering
        \includegraphics[width=\linewidth]{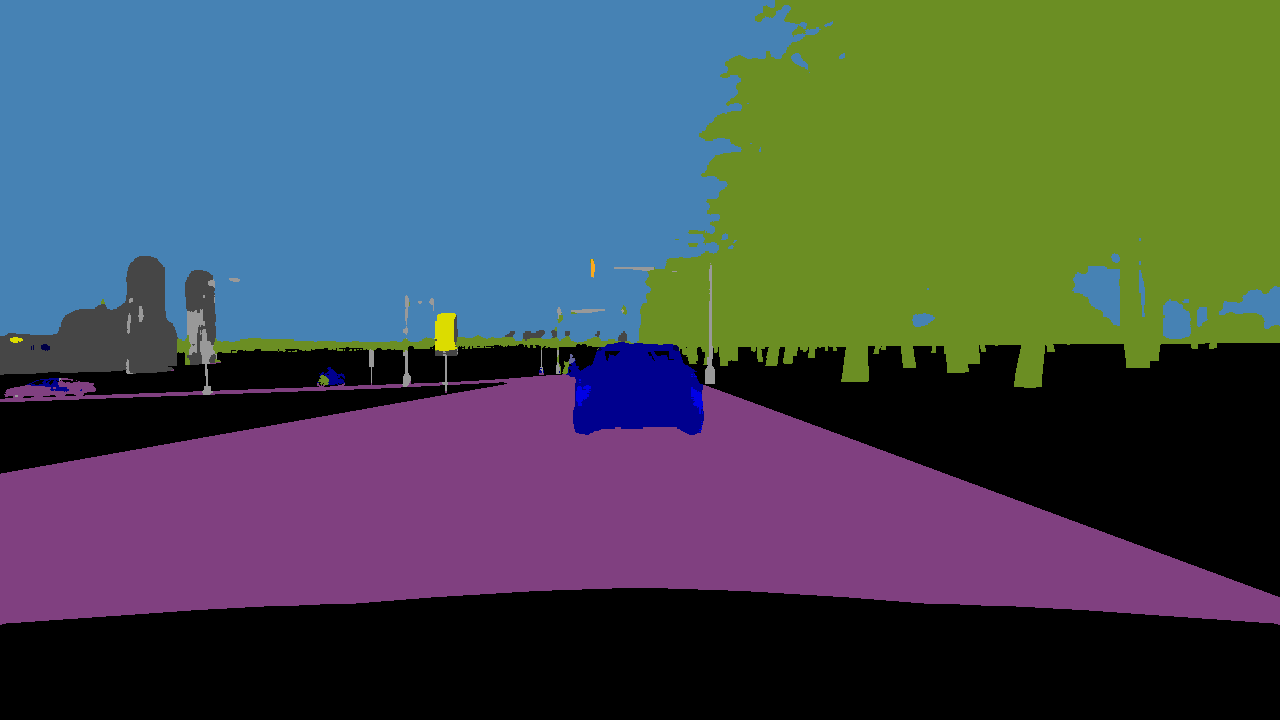}
    \end{subfigure}
    \begin{subfigure}{.17\linewidth}
        \centering
        \includegraphics[width=\linewidth]{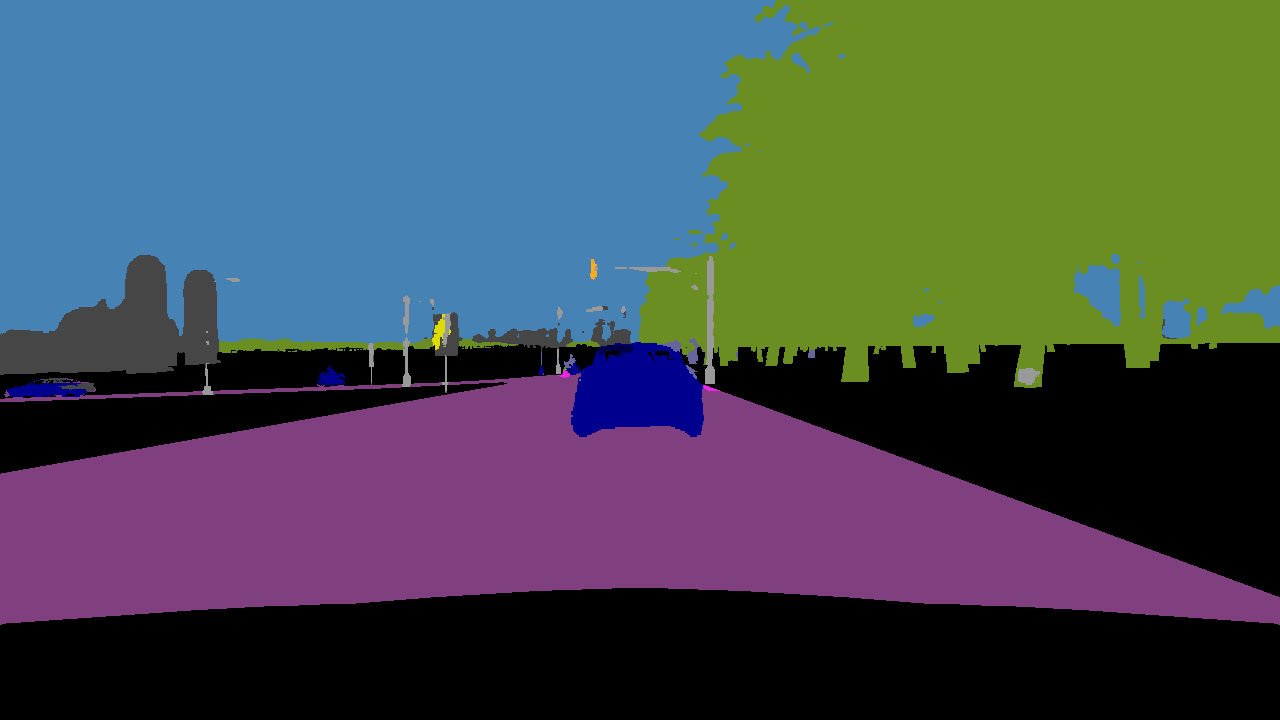}
    \end{subfigure}
    \begin{subfigure}{.17\linewidth}
        \centering
        \includegraphics[width=\linewidth]{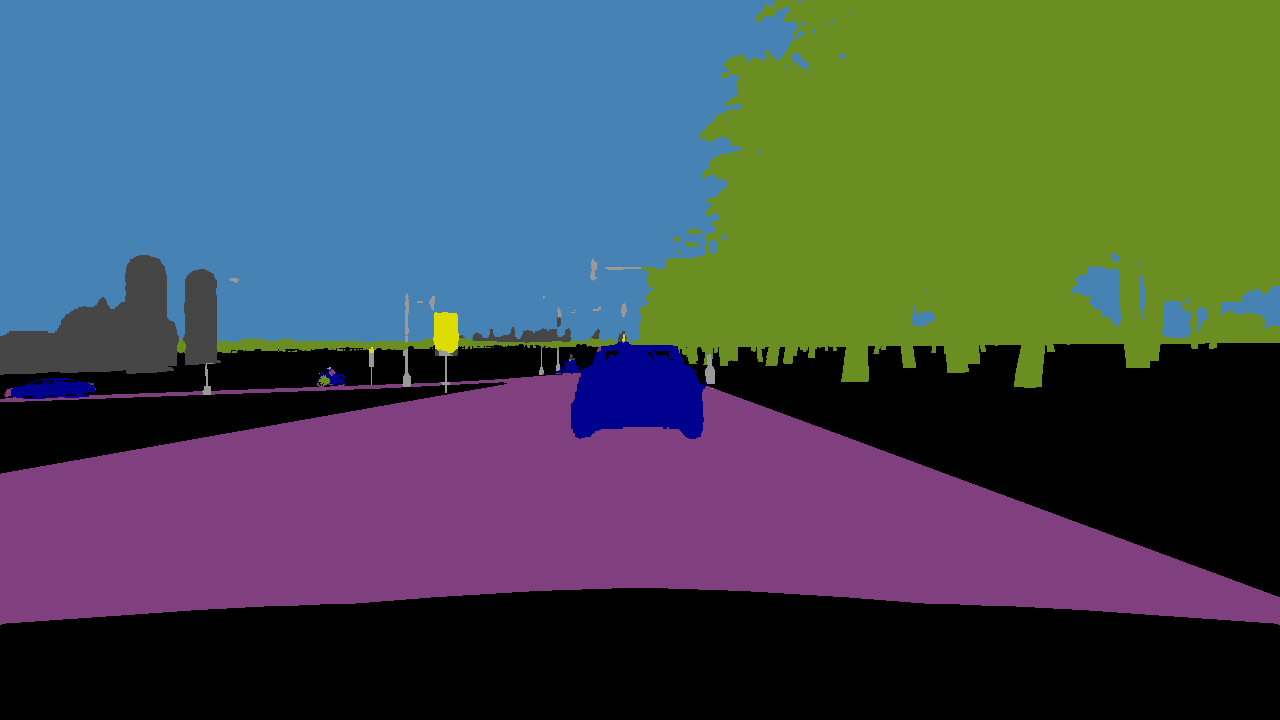}
    \end{subfigure}
    \begin{subfigure}{.17\linewidth}
        \centering
        \includegraphics[width=\linewidth]{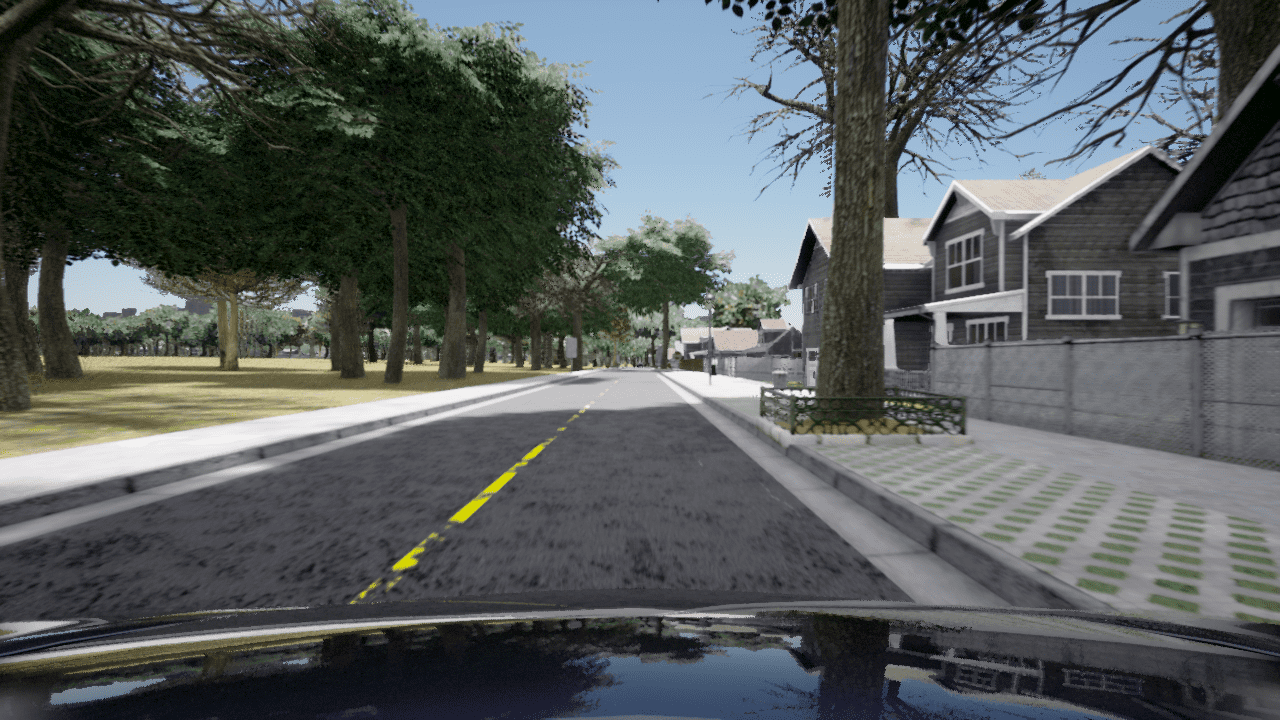}
    \end{subfigure}
    \begin{subfigure}{.17\linewidth}
        \centering
        \includegraphics[width=\linewidth]{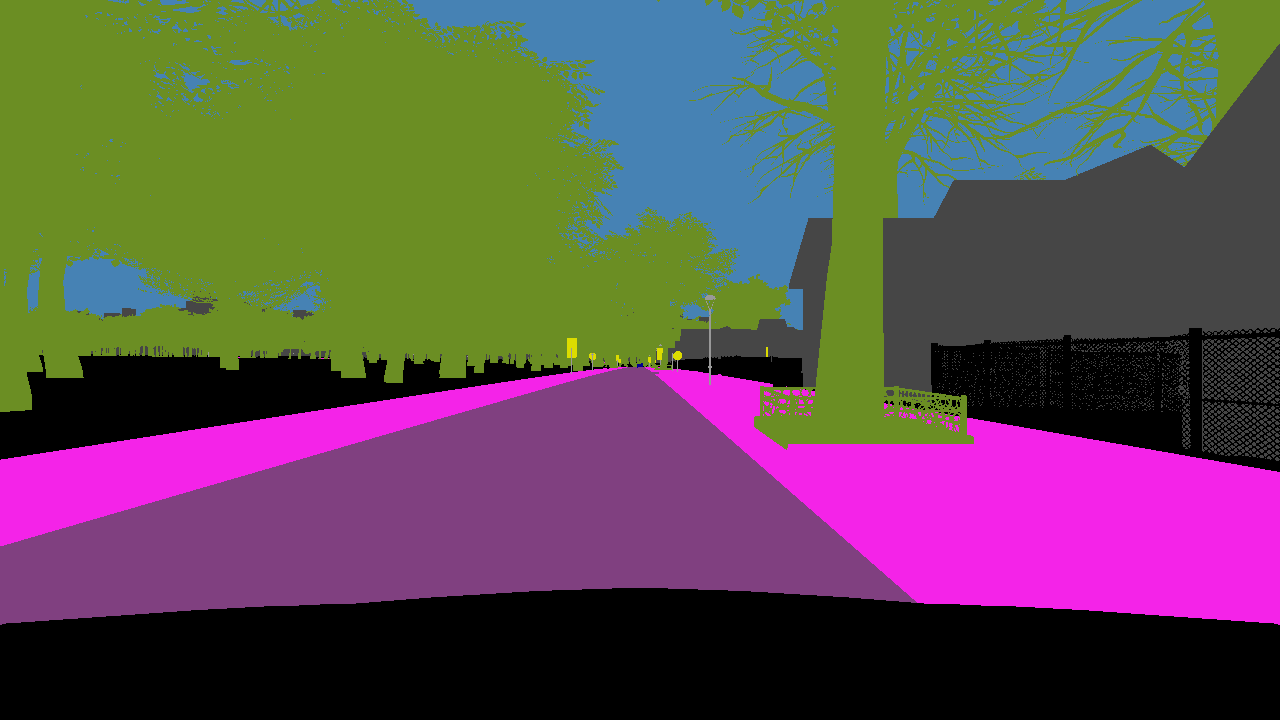}
    \end{subfigure}
    \begin{subfigure}{.17\linewidth}
        \centering
        \includegraphics[width=\linewidth]{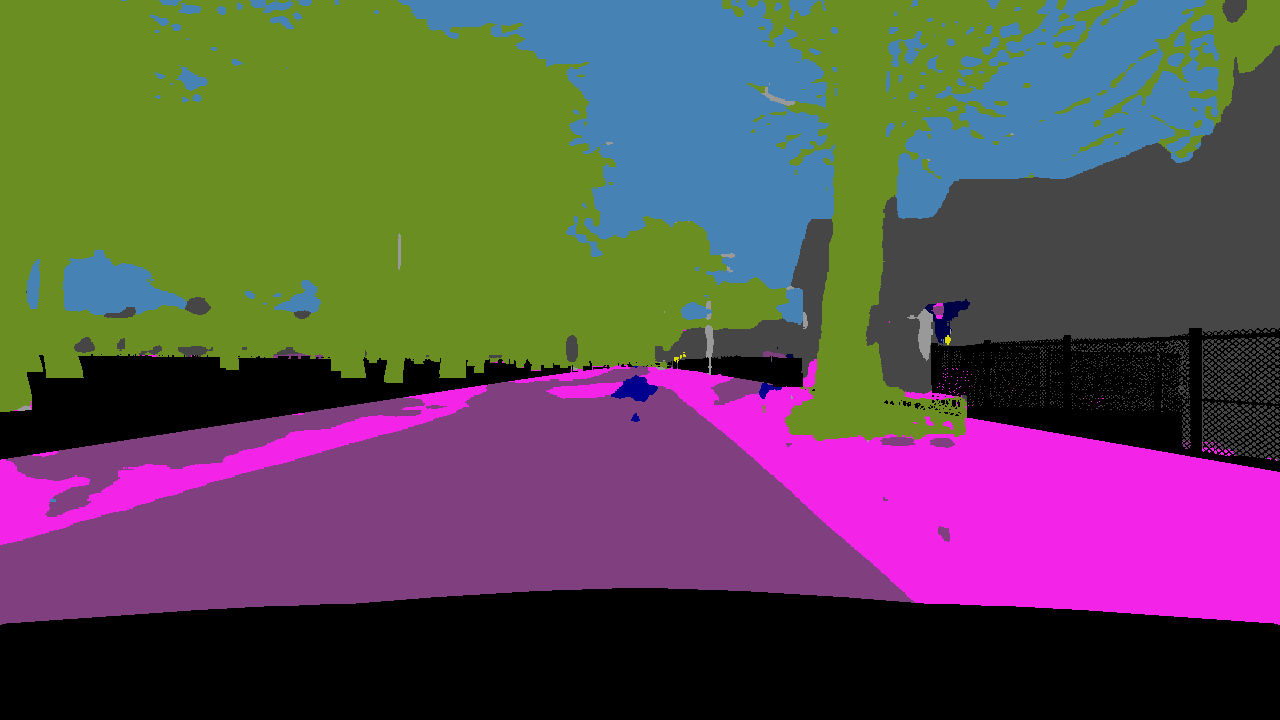}
    \end{subfigure}
    \begin{subfigure}{.17\linewidth}
        \centering
        \includegraphics[width=\linewidth]{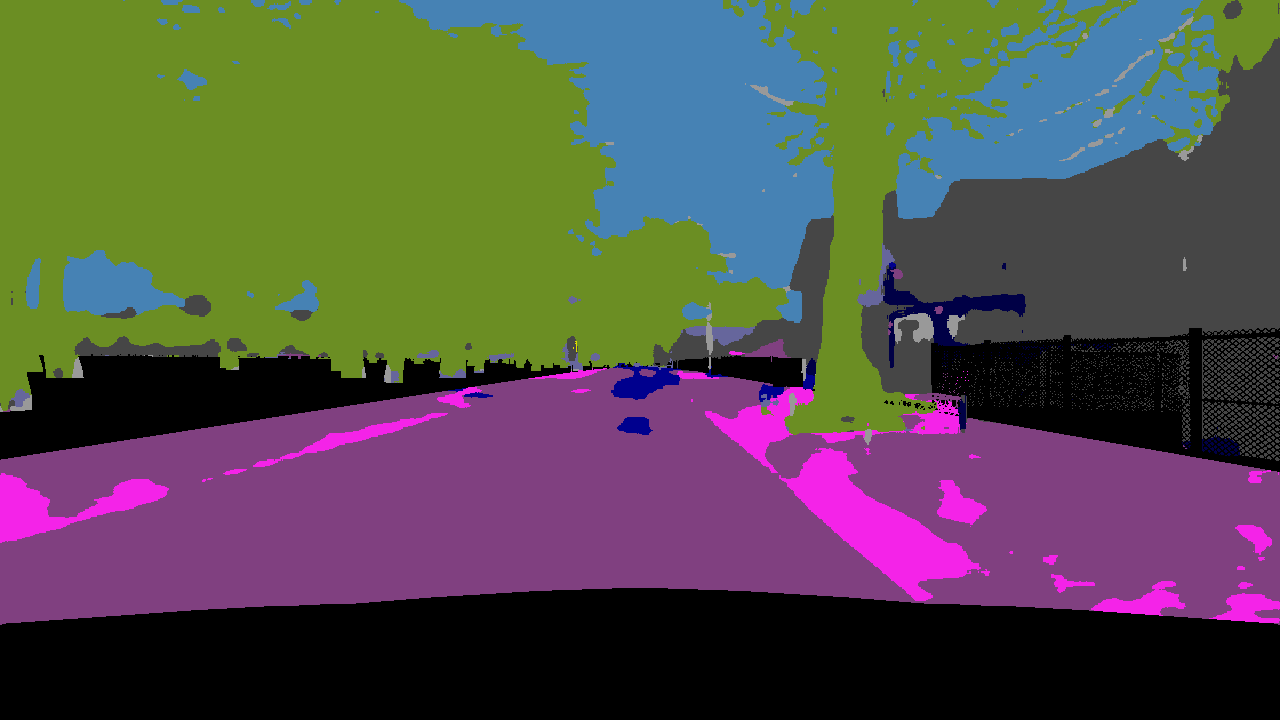}
    \end{subfigure}
    \begin{subfigure}{.17\linewidth}
        \centering
        \includegraphics[width=\linewidth]{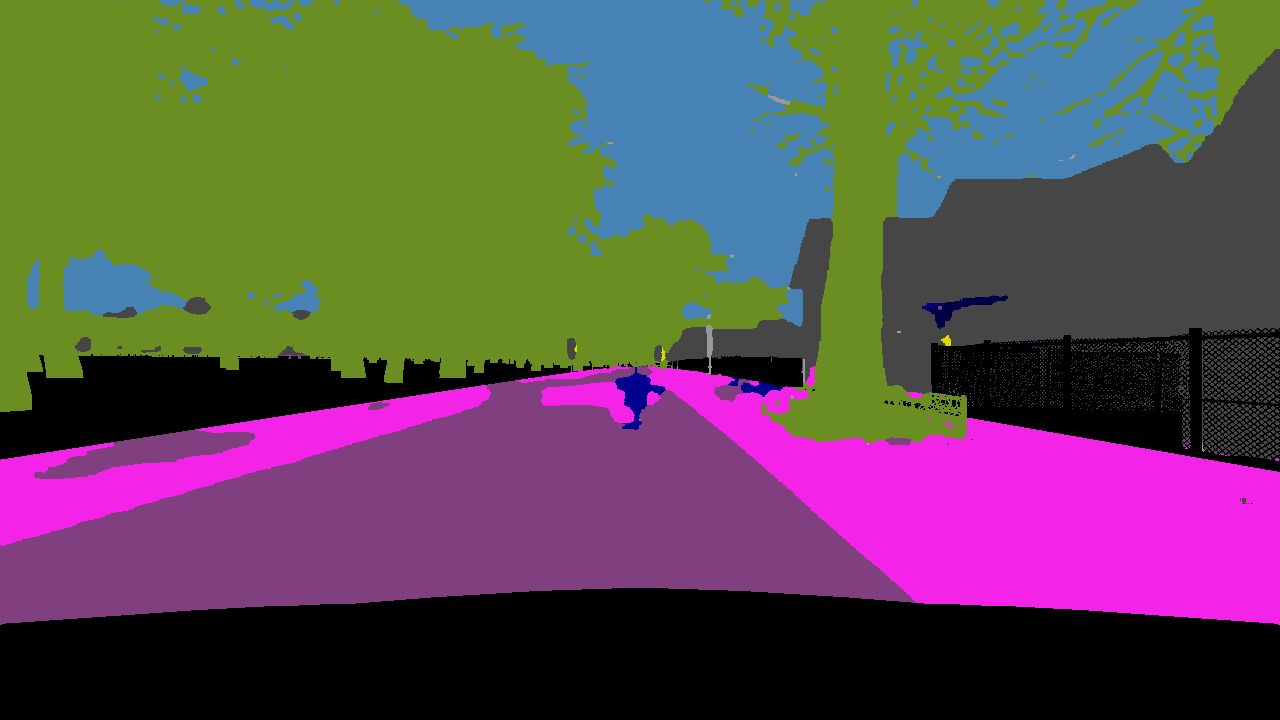}
    \end{subfigure}
    \caption{\textbf{Qualitative results.} Comparison of different segmentation masks. From left to right: RGB image, ground truth, \textit{Baseline}, \textit{Common} scenario with \textit{General} pretraining, and \textit{Spatial} scenario with \textit{Cell} pretraining. Black areas correspond to non-evaluated classes.}
    \label{fig:qualitatif}
\end{figure*}

\subsubsection{Quantitative performances}
Table~\ref{tab:agg_sw} shows the mean performance of the fleet
in the different settings on DADE-static. We observe that the baseline setups~\cite{Cioppa2019ARTHuS, Houyon2023Online} are outperformed by our method for every scenario and pretraining, highlighting the benefits of using multiple streams when adapting the models. For no pretraining (\textit{Scratch}), the \textit{Spatial} division of the environment leads to the best results, indicating that leveraging cellular information improves the models. For \textit{General} pretraining, the adapted \textit{Common} and \textit{Spatial} scenarios show better performance than the frozen pretrained model, highlighting the benefits of adapting the model online. 
In the MSC-TTA setup, the \textit{Cell} pretraining outperforms the \textit{General} one while it is the opposite in the MSC-OL setup, indicating that clean generic labels compensate for cell-specific ones.

We also provide the mean performance on DADE-dynamic in Table~\ref{tab:agg_dw}. As can be seen, our method still outperforms the baselines.
Again, from \textit{Scratch}, the \textit{Spatial} scenario brings the best results, followed by the \textit{Common} scenario. However, temporal divisions such as \textit{Weather}, \textit{Daylight}, and \textit{Specific} lead to lower performances. 
While DADE includes at least one vehicle in almost every location over time, the same weather and daylight are applied to all locations simultaneously, leading to discontinuities in the availability of samples for time-based cells. This temporarily stops the adaptation and slows down model convergence. Longer sequences, would allow the models to better explore those cells.
Finally, the \textit{Cell} pretraining shows the best overall performance for the MSC-OL/TTA setups, showing the advantage of dividing the environment into cells.

\subsubsection{Evolution of the fleet performances}
The evolution of the fleet performance over time is shown in Figure~\ref{fig:exp_sw}. For visualization purposes, we aggregate the confusion matrices in sliding windows of $8$ minutes to compute the mIoU. 
Regarding the static weather (top row), the \textit{Baseline} is outperformed by all settings of our method.
Interestingly, even if the baseline starts from pretrained weights and our method from scratch, we outperform the baseline in the MSC-OL setup and reach similar performances in the MSC-TTA setup.
Additionally, \textit{Cell} pretraining with the \textit{Spatial} scenario reaches the best performance in the MSC-TTA setup for the whole duration, keeping steady performance.
This is crucial for autonomous vehicles that need to operate similarly in all conditions.


We also show the performance for DADE-dynamic in Figure~\ref{fig:exp_sw} (bottom row). For both the MSC-OL and MSC-TTA setups, the \textit{Daylight} scenario with \textit{Cell} pretraining produces the best performance before nightfall, after which it drops while other scenarios, such as \textit{Spatial} with \textit{Cell} pretraining, become better options. This is due to the fact that the night models are not updated before nightfall while the location-based models are constantly updated, during day, dusk, and night. 
Nevertheless, it can be seen that the performance drops regardless of the scenario or pretraining during nightfall, leaving room for improvement in future works.

\subsubsection{Qualitative results}

We qualitatively show the improvement of our multi-stream cellular method over the ARTHuS~\cite{Cioppa2019ARTHuS} baseline. To do so, we display in Figure~\ref{fig:qualitatif} the segmentation masks predicted by our method in two scenarios: the \textit{Common} scenario with \textit{General} pretraining  and the \textit{Spatial} scenario with \textit{Cell} pretraining, and compare them to the masks predicted by the \textit{Baseline} and the ground truth labels. On the top row, we show a vehicle driving in the countryside under static (clear) weather at the end of the online training. We can see that the baseline confuses some building with poles and a car is misclassified as being part of the road, while our method is able to correctly segment it. The \textit{Spatial} model produces the most accurate segmentation masks as it is able to precisely segment the city and vegetation in the background and the cars on the left. On the bottom row, we show a vehicle driving in the low density residential location under static (clear) weather also at the end of the online training. As can be seen, the \textit{Common} model fails in this cell because it needs to learn a broader data distribution and looses accuracy due to its limited learning capacity. Contrarily, the \textit{Spatial} model is able to better learn that particular cell data distribution and therefore produces the best results.

\section{Conclusion}\label{sec:conclusion}

Our novel Multi-Stream Cellular Test-Time Adaptation (MSC-TTA) setup addresses multi-stream model adaptation in dynamic environments.
We focus on environments where data distribution shifts pose significant challenges. To do so, we divide the environments into cells, characterized by similar conditions such as location and weather.
Then, we propose a real-time method based on an adaptive student-teacher approach, leveraging the multiple streams and cellular information.
Experimental validation on autonomous vehicles illustrates the benefits of our MSC-TTA setup, showcasing better performance compared to a single-stream baseline. Our novel DADE dataset supports our experiments and provides a comprehensive benchmark for future studies in test-time adaptation of semantic segmentation models for autonomous vehicles.
This work represents a significant step forward in the field of test-time adaptation, holding promise for substantial contributions to IoT and autonomous driving.

\noindent\textbf{Acknowledgments.}
 B.~Gérin and A.~Halin are funded by the Walloon region under grant No.~2010235 (ARIAC by DIGITALWALLONIA4.AI). A.~Cioppa is funded by the F.R.S.-FNRS. M.~Henry is funded by PIT MecaTech under grant No.~C8650 (ReconnAIssance).
  The present research benefited from computational resources made available on Lucia, infrastructure funded by the Walloon Region under grant No.~1910247.

\FloatBarrier 


\appendix
\section{Supplementary Material}\label{app:supplementarymaterial}

In Section~\ref{sec:DADE_dataset}, we provide a description of our new dataset. The transition effects of agents moving between cells are briefly studied in Section~\ref{sec:analysis_transient_agent}. We also present additional experiments on cyclic domain shifts in Section~\ref{sec:other_datasets}.

\subsection{DADE Dataset}\label{sec:DADE_dataset}
To study our new Multi-Stream Cellular Test-Time Adaptation (MSC-TTA) setup and evaluate the performance of our real-time method, we need a dataset that meets the four following criteria. \textbf{(1) Multi-agent long videos:} the dataset should consist of long video sequences captured by multiple agents operating within the same dynamic environment.  \textbf{(2) Environment division:} the environment should be heterogeneous or dynamic to be spatially and/or temporally divided into cells, \eg, encompassing a variety of driving locations, such as rural, urban, and highway settings, or a broad spectrum of weather conditions, including, \eg, day, night, clear, rainy, and foggy scenarios. 
\textbf{(3) Cell connection:} each agent's connection to a cell should be precisely estimated, for instance using GNSS (Global Navigation Satellite System) coordinates for the location, or a weather service for the weather conditions. \textbf{(4) Available ground truths:} for evaluation purposes, we need to have access to ground-truth annotations for our semantic segmentation task.
Unfortunately, publicly-available datasets do not meet these criteria. Existing datasets, such as~\cite{Cordts2016The, Sun2022SHIFT, Maddern20171Year}, typically feature short video sequences, lack multi-agents, or often do not include ground-truth annotations or a diverse range of weather conditions.
While the SHIFT dataset~\cite{Sun2022SHIFT} contains varying weather conditions and ground truths, it is not a multi-agent dataset and its average sequence length is under $160$~s, which is too short for evaluating the long term impact of our method.

\begin{figure*}
    \centering
    \begin{subfigure}{.3\linewidth}
        \centering
        \includegraphics[width=\linewidth]{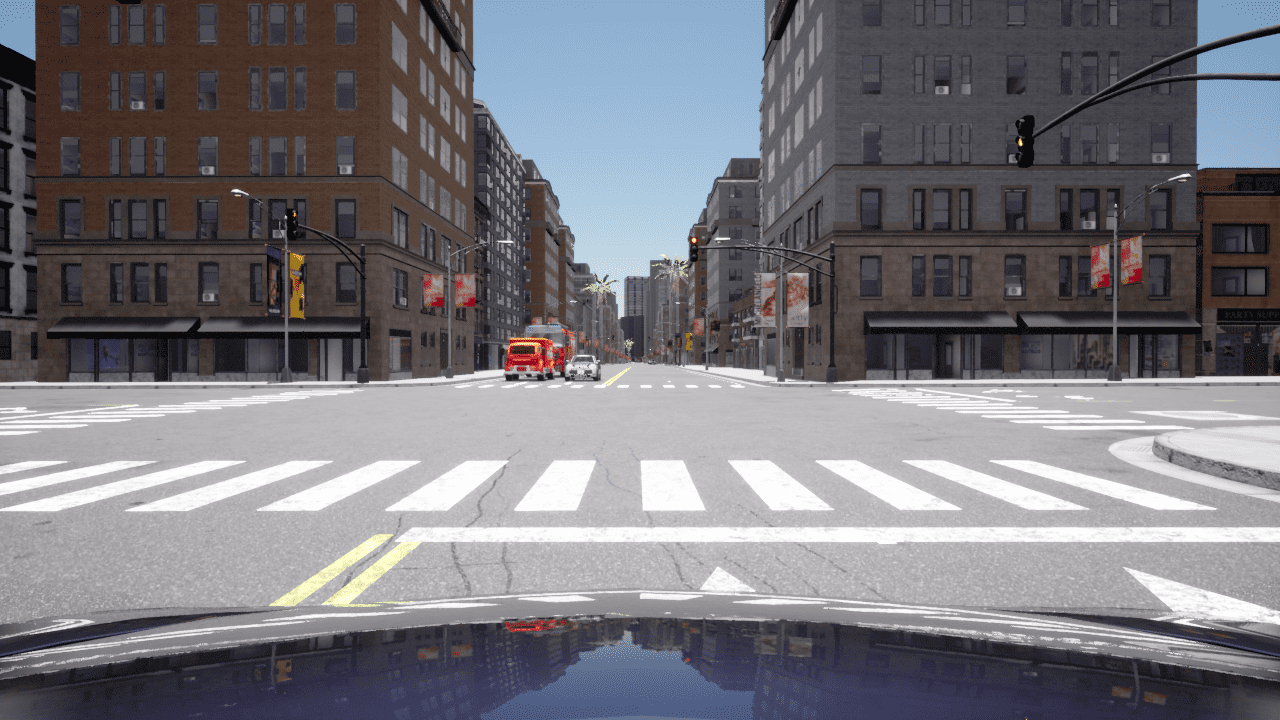}
        \caption{}
        \label{fig:semantic-masks-a}
    \end{subfigure}
    \hspace{1em}
    \begin{subfigure}{.3\linewidth}
        \centering
        \includegraphics[width=\linewidth]{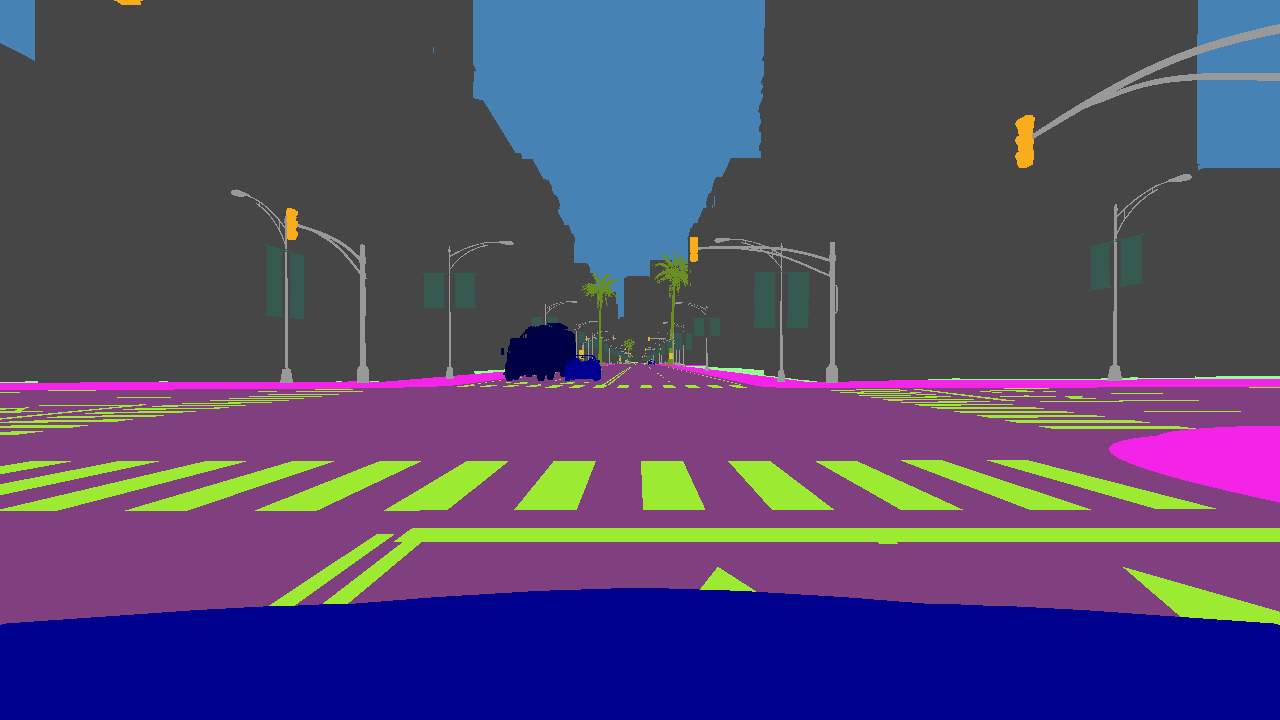}
        \caption{}
        \label{fig:semantic-masks-b}
    \end{subfigure}
    \hspace{1em}
    \begin{subfigure}{.3\linewidth}
        \centering
        \includegraphics[width=\linewidth]{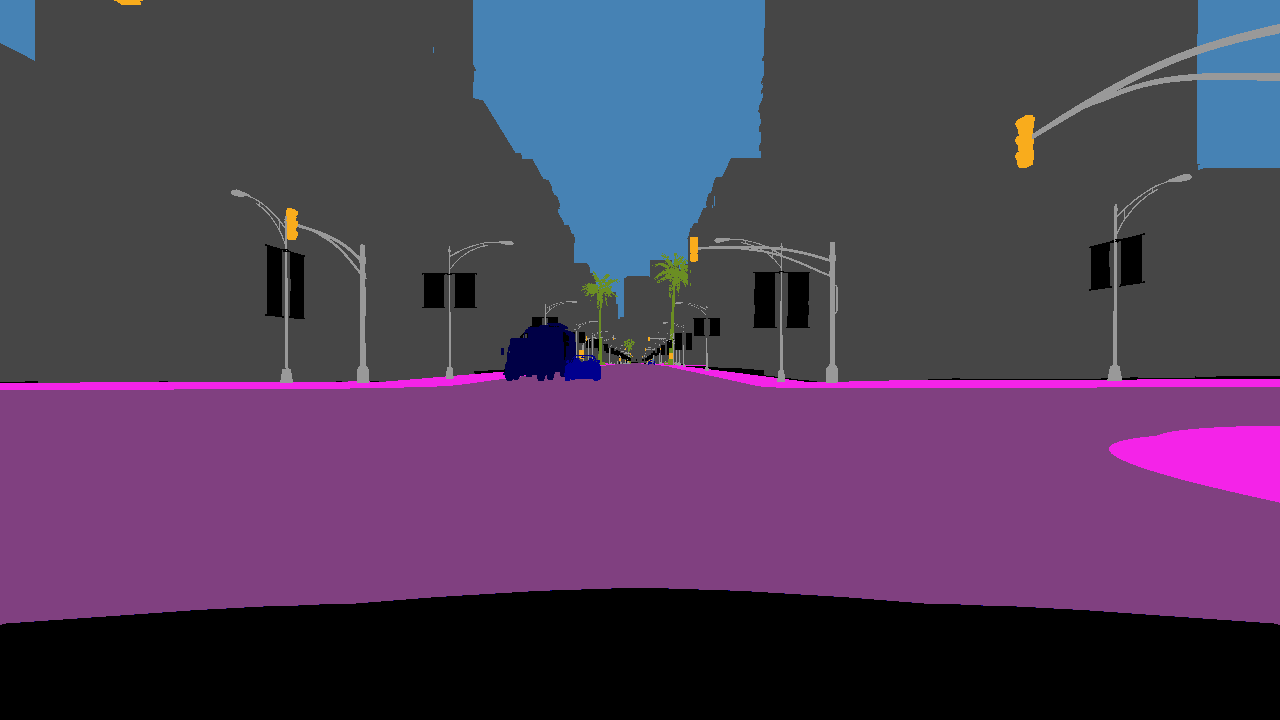}
        \caption{}
        \label{fig:semantic-masks-c}
    \end{subfigure}
    \caption{\textbf{Comparison between the real ground truth and the ground truth used in our experiments.} (a) An RGB image with (b) its corresponding semantic segmentation ground truth from CARLA and (c) the semantic segmentation ground truth that we used to evaluate our method. The black pixels in image (c) correspond to ignored classes or regions, such as the hood of the ego vehicle.}
    \label{fig:semantic-masks}
\end{figure*}

Therefore, we generated and will publicly release our own Driving Agents in Dynamic Environments (DADE) dataset, meeting all the above criteria. To have access to ground-truth annotations and precisely control the environment, we choose the CARLA simulator~\cite{Dosovitskiy2017CARLA} (version 0.9.14) to generate the dataset. We synchronize and calibrate all sensors and register the semantic segmentation ground truths. Our dataset is acquired using the recent \texttt{Town12} map that offers several visually distinct locations and fine-grained control over the weather.
Our simulation showcases several agents, in our case, ego vehicles, on which a camera is attached at the front, filming its front view (in a ``Cityscapes" fashion), as shown in Figure~\ref{fig:semantic-masks}.
We collect the video sequences taken by an RGB camera, the semantic segmentation ground-truth masks, the GNSS position of each agent in the simulation as well as the overall weather information. All signals are acquired at the framerate of $1$~frame per second, with a high-resolution (HD) definition. 
\begin{table}
    \centering
    \caption{\textbf{Comparison of class definition} between Cityscapes~\cite{Cordts2016The}, CARLA~\cite{Dosovitskiy2017CARLA}, and our DADE dataset. Our DADE dataset takes the intersection of the class definition between Cityscapes and CARLA. The classes not included in the intersection are projected to the ``unlabeled" class, except for ``road line" which is projected to ``road". The classes used in training and evaluation for DADE are the same as the ones of Cityscapes.}
    \resizebox{\columnwidth}{!}{
    \begin{tabular}{c|c|c||c||c|c|c}
        \multicolumn{3}{c||}{\textbf{Cityscapes}} & \multicolumn{1}{c||}{\textbf{CARLA}} & \multicolumn{3}{c}{\textbf{DADE}}\\ \midrule
        \textbf{name} & \textbf{training} & \textbf{evaluation} & \textbf{name} & \textbf{name} & \textbf{training} & \textbf{evaluation}\\ \midrule
        unlabeled & & & unlabeled & unlabeled & & \\
        static & & & static & static & & \\
        dynamic & & & dynamic & dynamic & & \\
        ground & & & ground & ground & & \\
        road & \checkmark & \checkmark & road & road & \checkmark & \checkmark \\
        sidewalk & \checkmark & \checkmark & sidewalk & sidewalk & \checkmark & \checkmark \\
        rail track & & & rail track & rail track & & \\
        building & \checkmark & \checkmark & building & building & \checkmark & \checkmark \\
        wall & \checkmark & \checkmark & wall & wall & \checkmark & \checkmark \\
        fence & \checkmark & \checkmark & fence & fence & \checkmark & \checkmark \\
        guard rail & & & guard rail & guard rail & & \\
        bridge & & & bridge & bridge & & \\
        pole & \checkmark & \checkmark & pole & pole & \checkmark & \checkmark \\
        traffic light & \checkmark & \checkmark & traffic light & traffic light & \checkmark & \checkmark \\
        traffic sign & \checkmark & \checkmark & traffic sign & traffic sign & \checkmark & \checkmark \\
        vegetation & \checkmark & \checkmark & vegetation & vegetation & \checkmark & \checkmark \\
        terrain & \checkmark & \checkmark & terrain & terrain & \checkmark & \checkmark \\
        sky & \checkmark & \checkmark & sky & sky & \checkmark & \checkmark \\
        person & \checkmark & \checkmark & person & person & \checkmark & \checkmark \\
        rider & \checkmark & \checkmark & rider & rider & \checkmark & \checkmark \\
        car & \checkmark & \checkmark & car & car & \checkmark & \checkmark \\
        truck & \checkmark & \checkmark & truck & truck & \checkmark & \checkmark \\
        bus & \checkmark & \checkmark & bus & bus & \checkmark & \checkmark \\
        motorcycle & \checkmark & \checkmark & motorcycle & motorcycle & \checkmark & \checkmark \\
        bicycle & \checkmark & \checkmark & bicycle & bicycle & \checkmark & \checkmark \\
        ego vehicle & & & ego vehicle & ego vehicle & & \\
        rectification border & & & other &  & & \\
        out of roi & & & road line &  & & \\
        parking & & & water & & & \\
        tunnel & & & & & & \\
        caravan & & & & & & \\
        trailer & & & & & & \\
        train & & & & & & \\ \bottomrule
        
    \end{tabular}}
    \label{tab:class_def}
\end{table}

To align our dataset with current benchmarks in the semantic segmentation field, we generated two versions of the semantic segmentation ground truths in the dataset: (1) the ones directly collected from the CARLA simulator (including $29$ semantic classes), and (2) the intersection of the semantic classes available in CARLA and the semantic classes from the Cityscapes~\cite{Cordts2016The} dataset (including $33$ different semantic classes).
For consistency with previous works, we choose the later version in our experiments, as most state-of-the-art models for semantic segmentation in driving environments are trained on Cityscapes. 
Nevertheless, the discrepancies between the two versions are minimal and could be interchanged depending on the user's preferences. Table~\ref{tab:class_def} provides the complete comparison between the semantic segmentation classes of the Cityscapes dataset, the CARLA simulator and our DADE dataset.
Figure~\ref{fig:semantic-masks-a} shows an RGB image alongside the two versions of the semantic segmentation ground-truth masks (Figure~\ref{fig:semantic-masks-b} and Figure~\ref{fig:semantic-masks-c}). As can be seen, the ``road line" class in the CARLA labels, visible in Figure~\ref{fig:semantic-masks-b}, does not exist in the Cityscapes labels. Also, the ``car hood" is ignored (indicated by black pixels) in the second version.

\begin{figure}
    \centering
    \includegraphics[width=0.85\linewidth]{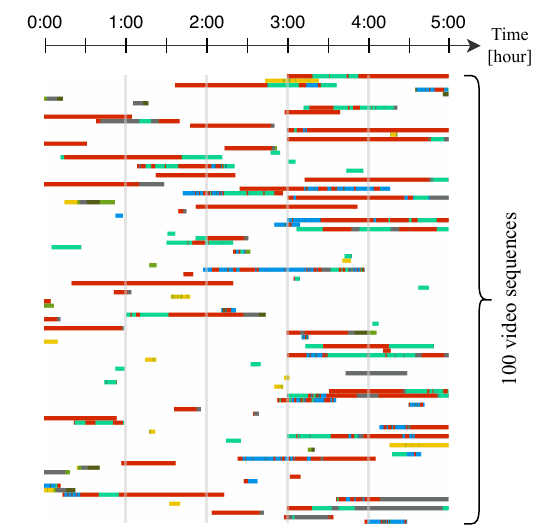}
    \caption{\textbf{Location of each agent} for the $100$ sequences of the DADE-static dataset. The color of the line corresponds to the location of the agent for each sequence at a given time: \textcolor{forest}{forest}, \textcolor{countryside}{countryside}, \textcolor{rural_farmland}{rural farmland}, \textcolor{highway}{highway}, \textcolor{low_density_residential}{low density residential}, \textcolor{community_buildings}{community buildings}, and \textcolor{high_density_residential}{high density residential}. We can see that the sequences are evenly distributed across the entire 5-hour time frame.}
    \label{fig:fixed-weather-sequences}
\end{figure}

\begin{figure}
    \centering
    \includegraphics[width=0.85\linewidth]{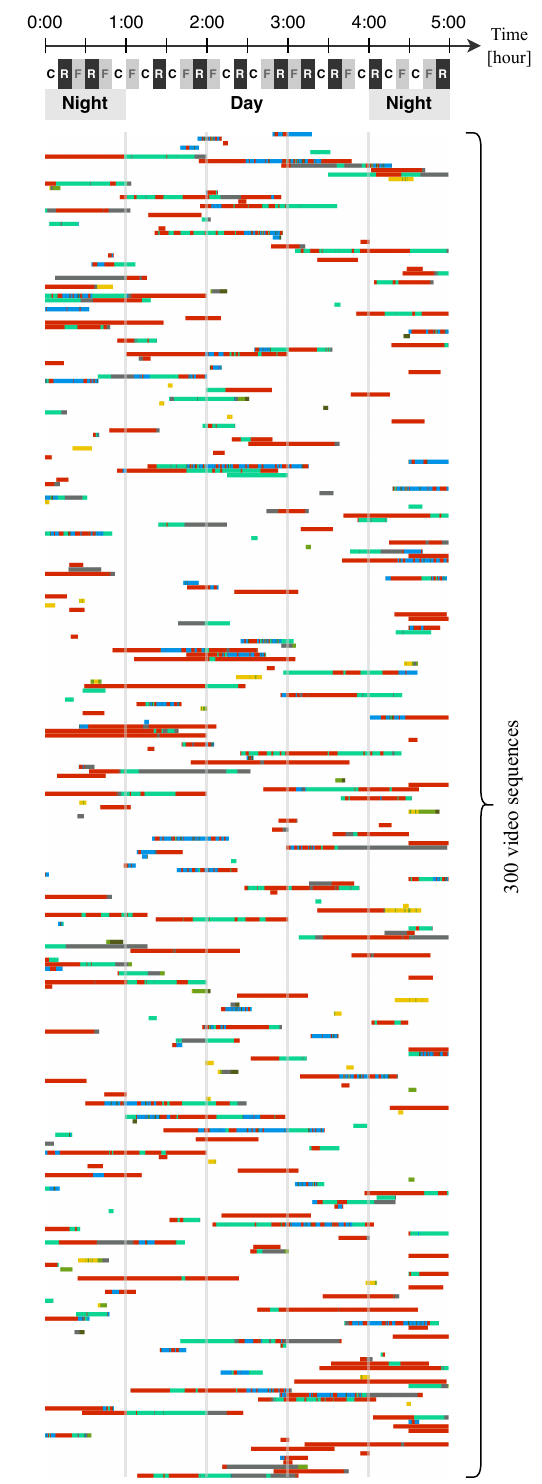}
    \caption{\textbf{Location, weather, and daylight conditions of each agent} for the $300$ sequences of the DADE-dynamic dataset. C, R and F respectively correspond to clear, rainy and foggy weathers, and night/day represent the daylight conditions. The color of the line corresponds to the location of the agent for each sequence at a given time: \textcolor{forest}{forest}, \textcolor{countryside}{countryside}, \textcolor{rural_farmland}{rural farmland}, \textcolor{highway}{highway}, \textcolor{low_density_residential}{low density residential}, \textcolor{community_buildings}{community buildings}, and \textcolor{high_density_residential}{high density residential}.}
    \label{fig:dynamic-weather-sequences}
\end{figure}

\begin{figure}
    \centering
    \includegraphics[width=\linewidth]{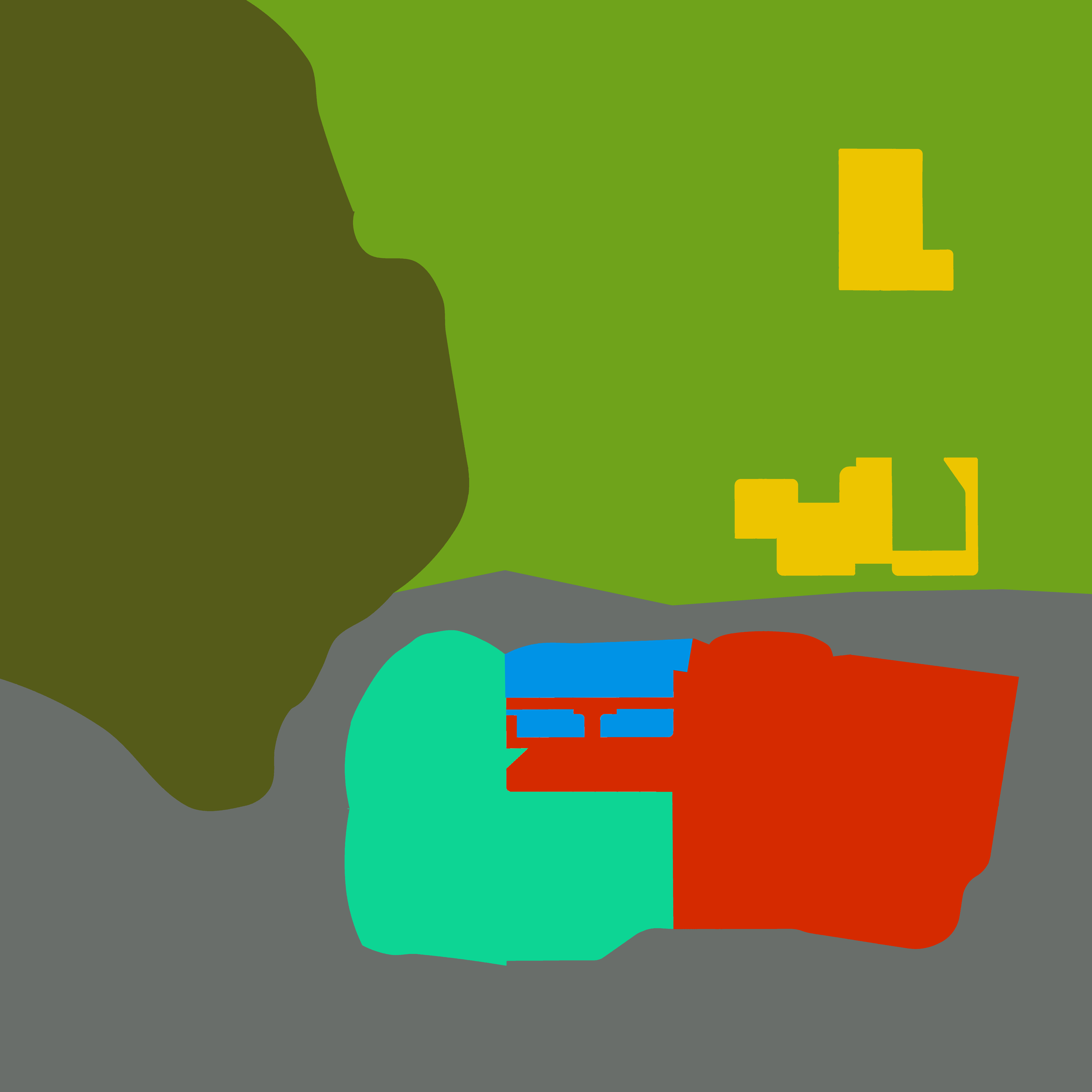}
    \caption{\textbf{Locations of the \texttt{Town12} map in CARLA.} The figure provides the location in which the vehicle is depending on its x and y coordinates. The different locations are the following: \textcolor{forest}{forest}, \textcolor{countryside}{countryside}, \textcolor{rural_farmland}{rural farmland}, \textcolor{highway}{highway}, \textcolor{low_density_residential}{low density residential}, \textcolor{community_buildings}{community buildings}, and \textcolor{high_density_residential}{high density residential}.}
    \label{fig:zones_town12}
\end{figure}

\begin{figure*}
    \centering
    \includegraphics[height=0.9\textheight]{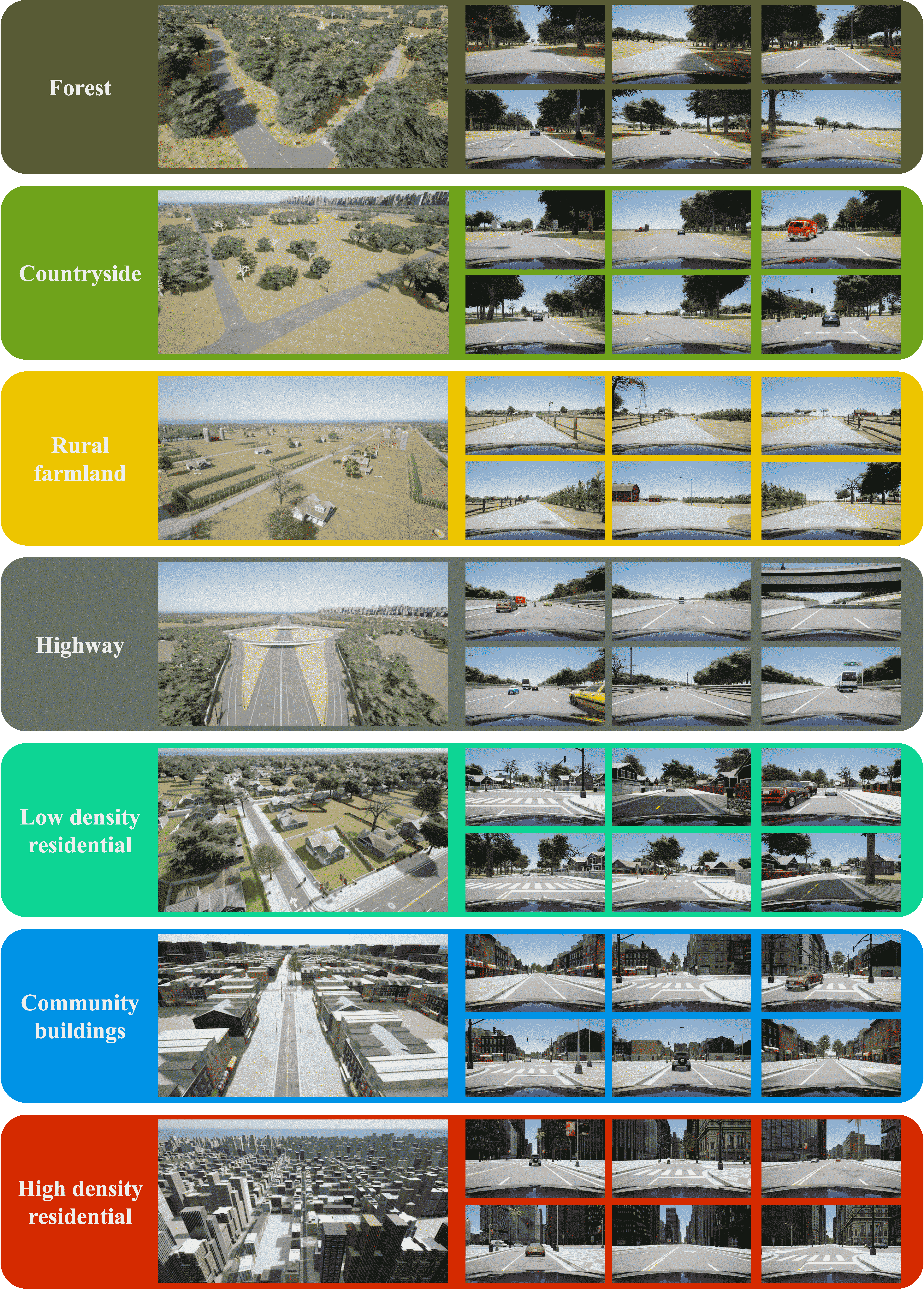}
    \caption{\textbf{Examples of the different locations in our dataset.} We define $7$ different locations based on the GNSS data and show some images captured by the agent in each location. From left to right, we display the name of the location, an overview of the location, and six images captured by agents.}
    \label{fig:DADE-zones}
\end{figure*}

\begin{figure}
    \centering
    \includegraphics[width=\linewidth]{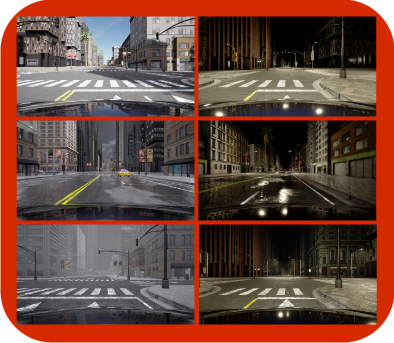}
    \caption{\textbf{Examples of the different weather and daylight conditions in our dataset.} The images show the $6$ different weather and daylight conditions in the same \textcolor{high_density_residential}{high density residential} location. The images correspond respectively, from left to right and top to bottom, to clear day, clear night, rainy day, rainy night, foggy day, and foggy night.}
    \label{fig:DADE-dynamic-weather-conditions}
\end{figure}

To study different cell divisions of the environment, our DADE dataset is composed of two parts. The first part, \emph{DADE-static}, is acquired with static weather conditions (clear day) and contains $100$ video sequences, as shown in Figure~\ref{fig:fixed-weather-sequences}. The second part, \emph{DADE-dynamic}, is acquired with varying weather conditions (ranging from day to night, with clear, rainy or foggy conditions) and contains $300$ video sequences, as shown in Figure~\ref{fig:dynamic-weather-sequences}.
For both parts, each sequence is acquired by one agent (one ego vehicle) running for some time within a $5$-hour time frame, amounting to a total of $990$k frames for the entire dataset. 
In Figure~\ref{fig:zones_town12}, we show a top view of the various locations in the \texttt{Town12} map of the CARLA simulator in which the agents evolve, namely \textcolor{forest}{forest}, \textcolor{countryside}{countryside}, \textcolor{rural_farmland}{rural farmland}, \textcolor{highway}{highway}, \textcolor{low_density_residential}{low density residential}, \textcolor{community_buildings}{community buildings}, and \textcolor{high_density_residential}{high density residential}. Images captured in each location can be seen in Figure~\ref{fig:DADE-zones}.
Finally, Figure~\ref{fig:DADE-dynamic-weather-conditions} illustrates the $6$ different weather conditions, in the \textcolor{high_density_residential}{high density residential} location, encountered in the DADE-dynamic dataset. 

Let us note that due to the limitations of the CARLA simulator running the \texttt{Town12} map, there are no pedestrians on the streets, only vehicles such as cars, motorcycles, bicycles or trucks. Also, the quantity of vehicles (traffic) is independent on the location. The vehicles spawned in the map move randomly through the seven locations. Finally, the different sequences are collected sequentially. In the following, we provide some statistics about both parts of our DADE dataset.

\subsubsection{DADE-static dataset}
\begin{figure*}
    \centering
    \begin{subfigure}{.48\linewidth}
        \centering
        \includegraphics[width=\linewidth]{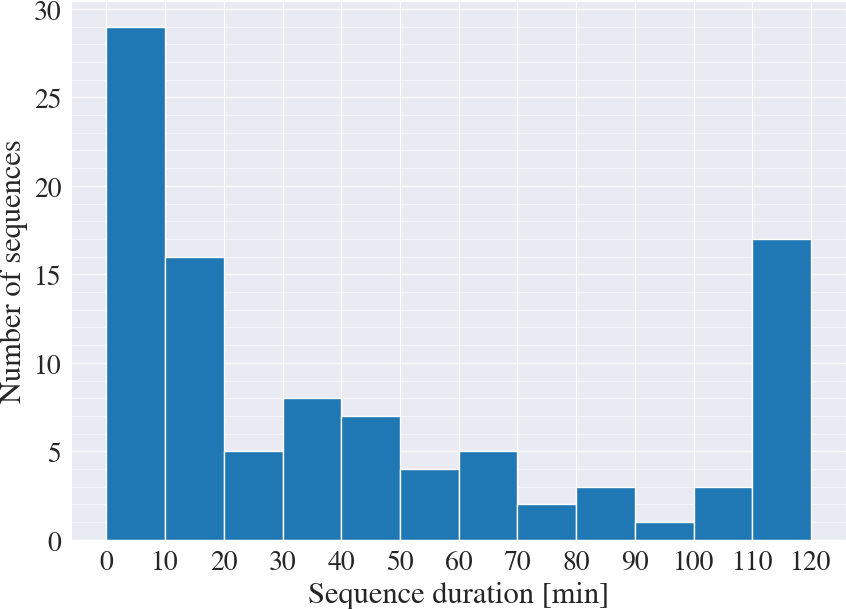}
        \caption{DADE-static}
        \label{fig:seq_durations-static}
    \end{subfigure}
    \hspace{1em}
    \begin{subfigure}{.48\linewidth}
        \centering
        \includegraphics[width=\linewidth]{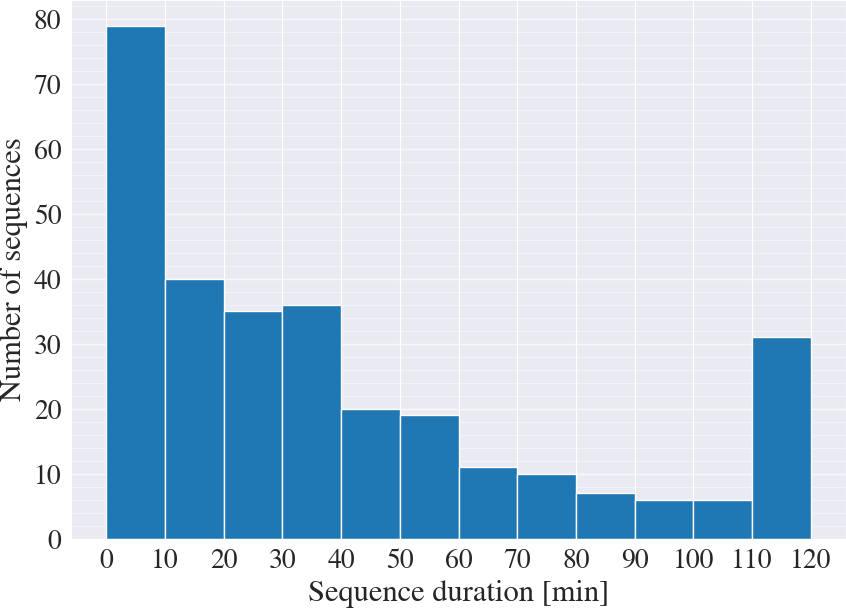}
        \caption{DADE-dynamic}
        \label{fig:seq_durations-dynamic}
    \end{subfigure}
    \caption{\textbf{Distribution of sequence lengths} for (a) the DADE-static dataset and (b) the DADE-dynamic dataset. The DADE-static and DADE-dynamic datasets have respectively an average sequence length of $45$ and $40$ minutes, with durations ranging from a few minutes to two hours.}
    \label{fig:seq_durations}
\end{figure*}

This first part of our dataset is composed of $100$ sequences, acquired in the \texttt{Town12} map of CARLA with a static clear sunny weather during the day. Each sequence contains between $271$ and $7{,}200$ frames acquired at $1$ fps, for total of $270{,}527$ frames, amounting to more than $75$ hours of video. In Figure~\ref{fig:seq_durations-static}, we show the distribution of the sequence length for the $100$ sequences. As can be seen, our dataset contains a lot of short and long sequences, with an average sequence length of $45$ minutes.
We also show, in Figure~\ref{fig:fixed-weather-sequences}, the locations of the $100$ agents over time. The colors correspond to the locations in which the agents are evolving (see Figure~\ref{fig:zones_town12}). We can see that, for most sequences, the agents evolve through several locations, and that the start and end times vary significantly from one agent to another. 

\begin{figure*}
    \centering
    \begin{subfigure}{.48\linewidth}
        \centering
        \includegraphics[width=\linewidth]{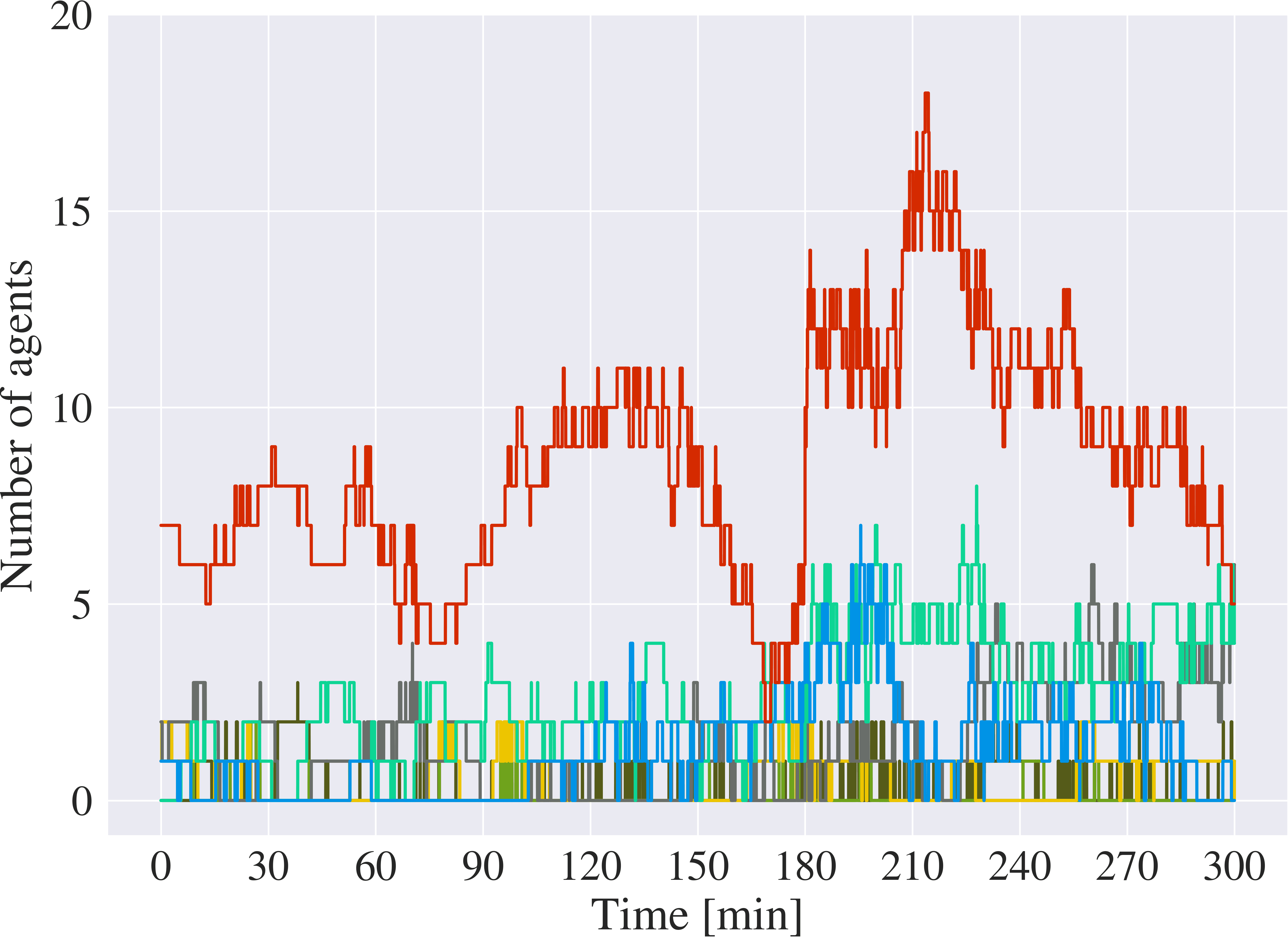}
        \caption{DADE-static}
        \label{fig:nb_vehicles-static}
    \end{subfigure}
    \hspace{1em}
    \begin{subfigure}{.48\linewidth}
        \centering
        \includegraphics[width=\linewidth]{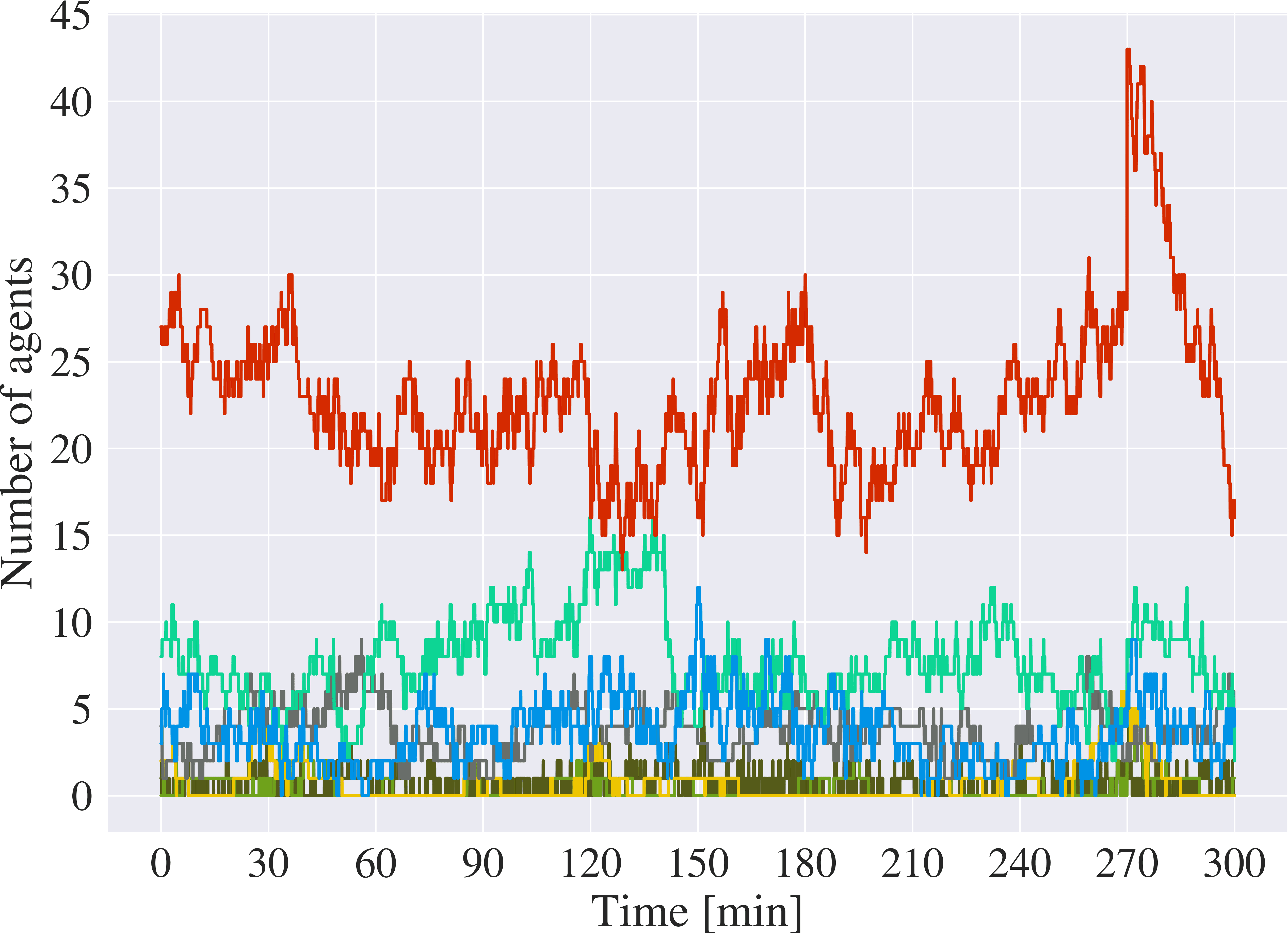}
        \caption{DADE-dynamic}
        \label{fig:nb_vehicles-dynamic}
    \end{subfigure}
    \caption{\textbf{Number of agents per location} over time for (a) the DADE-static dataset and (b) the DADE-dynamic dataset. The colors of the plots correspond to the location: \textcolor{forest}{forest}, \textcolor{countryside}{countryside}, \textcolor{rural_farmland}{rural farmland}, \textcolor{highway}{highway}, \textcolor{low_density_residential}{low density residential}, \textcolor{community_buildings}{community buildings}, and \textcolor{high_density_residential}{high density residential}. The same trends can be observed in both datasets, with three times as many agents in DADE-dynamic as in DADE-static. Note that there is at all time at least one agent in the high density residential location in both dataset and in the low density residential in the DADE-dynamic. Conversely, forest, countryside, and rural farmland locations exhibit the least agent presence, often remaining empty of agents for extended periods.}
    \label{fig:nb_vehicles}
\end{figure*}

Figure~\ref{fig:nb_vehicles-static} provides a more detailed analysis of each agent's location over time. 
Particularly, it shows that there is a high imbalance between the locations, which is expected in real-world scenarios. For instance, it is realistic to encounter much more vehicles in city centers than in the countryside.
Table~\ref{tab:Nb_images_fixed_weather} summarizes those values and splits the number of images acquired during the two first hours (used for pretraining) and the three last hours (used for testing). Interestingly, data originating from the high density residential location constitute over half of our DADE-static dataset. We can also see that, during the first two hours, over a thousand images are collected in each location, constituting a sufficient pretraining set. 

\begin{table}[ht]
\centering
\caption{\textbf{Number of images per location} within the DADE-static dataset during the pretraining time (two first hours), the test time (three last hours), and the overall time (the five hours), as well as the proportion of images originating from each location in comparison to the entire dataset.}
\resizebox{\columnwidth}{!}{
\begin{tabular}{l|c|c|c|c}
                         &  \textbf{Pretraining} &  \textbf{Testing}    &  \textbf{Overall}       &  \textbf{Proportion of}       \\ 
\textbf{Location}            &  \textbf{(2 hours)}   &  \textbf{(3 hours)}  &  \textbf{(5 hours)}   &  \textbf{the entire dataset}   \\ \hline
Forest                   & 2{,}176                  & 2{,}796                 & 4{,}972                  &  1.84\%                       \\ \hline
Countryside              & 2{,}442                  & 1{,}215                 & 3{,}657                  &  1.35\%                       \\ \hline
Rural farmland           & 3{,}608                  & 6{,}089                 & 9{,}697                  &  3.58\%                       \\ \hline
Highway                  & 7{,}018                  & 19{,}159                & 26{,}177                 &  9.68\%                       \\ \hline
Low density residential  & 11{,}187                 & 36{,}658                & 47{,}845                 &  17.69\%                      \\ \hline
Community buildings      & 2{,}357                  & 20{,}404                & 22{,}761                 &  8.41\%                       \\ \hline
High density residential & 50{,}034                 & 105{,}384               & 155{,}418                &  57.45\%                       \\ \hline
\textbf{Total}           & \textbf{78{,}822}        & \textbf{191{,}705}      & \textbf{270{,}527}       &  \textbf{100\%}     
\end{tabular}}

\label{tab:Nb_images_fixed_weather}
\end{table}

\subsubsection{DADE-dynamic dataset}
\begin{figure*}
    \centering
    \begin{subfigure}{.48\linewidth}
        \centering
        \includegraphics[width=\linewidth]{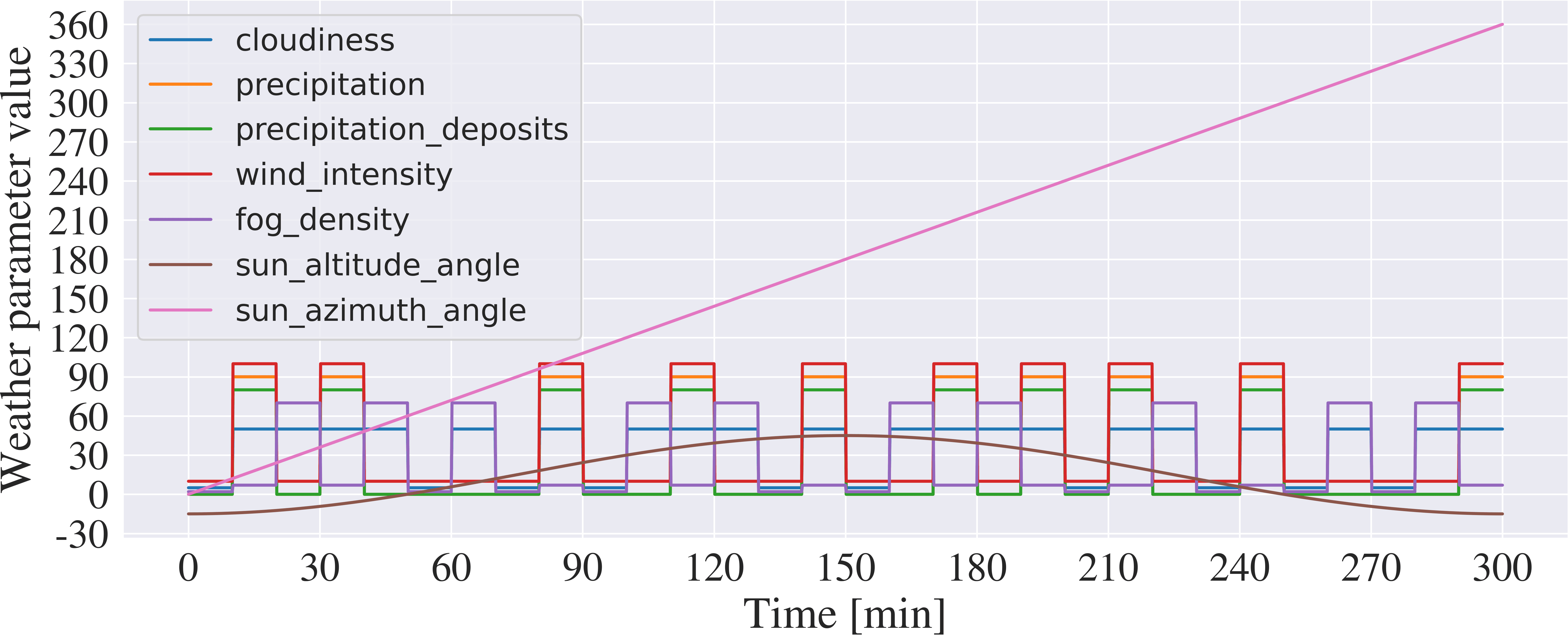}
        \caption{}
        \label{fig:weather_parameters}
    \end{subfigure}
    \hspace{1em}
    \begin{subfigure}{.48\linewidth}
        \centering
        \includegraphics[width=\linewidth]{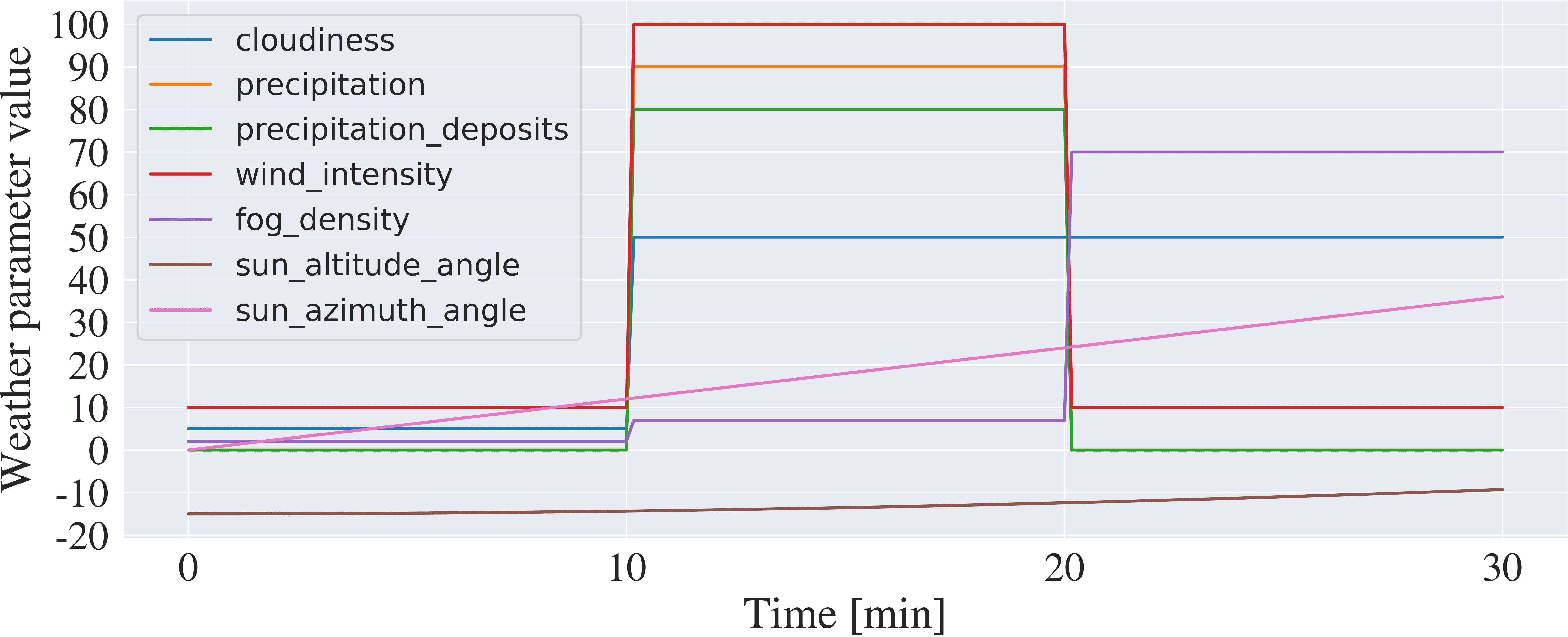}
        \caption{}
        \label{fig:weather_parameters_zoom}
    \end{subfigure}
    \caption{\textbf{Evolution of the weather parameters over time.} The weather switches arbitrarily every $10$ minutes between clear, rainy, and foggy weathers with a smooth transition of $10$ seconds. The parameters for the daylight conditions are related to the sun position, \ie sun altitude angle and sun azimuth angle, which vary smoothly over time, respectively between $-15$ and $45$ degrees, and between $0$ and $360$ degrees. We consider that it is night time during the first hour and the last hour, when the sun altitude is below $5$ degrees, and that it is the day in between during three hours, \ie, when the sun's altitude is over $5$ degrees. In total, the clear, rainy, and foggy weathers each occur $10$ times; $4$ times during the night and $6$ times during the day, as shown in (a). (b) zooms in on the first thirty minutes. During the first ten minutes, the weather is clear, then there is a smooth transition of $10$ seconds towards a rainy weather and, finally, after $20$ minutes there is again a smooth transition of $10$ seconds towards a foggy weather.}
    \label{fig:weather_parameters_with_zoom}
\end{figure*}

This second part of our dataset is acquired during a period of time of $5$ hours with varying weather conditions as shown in Figure~\ref{fig:dynamic-weather-sequences}. Particularly, it is composed of $300$ sequences containing between $188$ and $7{,}200$ frames acquired at $1$ fps, for a total of $719{,}742$ frames or $200$ hours of videos. Figure~\ref{fig:seq_durations-dynamic} shows the distribution of the sequence length. It can be noted that the distribution follows the same trend as for DADE-static, with a similar average sequence length of $40$ minutes.
To generate various weather conditions, we dynamically change the weather parameters over time in the same way for the entire map, \ie, that the weather condition is the same for all agents at a given time. We change the weather condition every $10$ minutes arbitrarily between clear, rainy, and foggy weathers, with a smooth transition on the weather parameters during $10$ seconds. For the daylight conditions, we choose to start the simulation during night time and let the sun rise after one hour, finally setting in the last hour. The $5$ hours are thus composed of a total of approximately $2$ hours of night conditions and $3$ hours of day conditions. During the entire simulation, there are $4$ periods of $10$ minutes for each weather condition (\ie clear, rainy, and foggy) during the night and $6$ during the day as shown in Figure~\ref{fig:weather_parameters}. To visualize the transitions, Figure~\ref{fig:weather_parameters_zoom} zooms in on the first thirty minutes where the simulation changes from clear, then rainy, and finally foggy weathers. 

Table~\ref{tab:Nb_images_dynamic_weather} provides a summary of the number of images for each location, weather, and daylight conditions during the two first hours (used for pretraining) and the three last hours (used for testing). We can see that the proportion of images in each location is similar to the one of our DADE-static dataset. However, upon further division based on both weather and daylight conditions, we can see a significant decrease in the number of images for each cell. Notably, this division results in the absence of pretraining data for the clear day weather condition in the countryside location.
Finally, Figure~\ref{fig:nb_vehicles-dynamic} shows the number of agents in each location. The color of the plots corresponds to the color code of the location in the \texttt{Town12} map (see Figure~\ref{fig:zones_town12}). As can be seen, there is also a high imbalance between the different locations.

\begin{table}[th]
\centering
\caption{\textbf{Number of images per location} within the DADE-dynamic dataset during the pretraining time (two first hours), the test time (three last hours), and the overall time (the five hours), as well as the proportion of images originating from each location in comparison to the entire dataset.}
\resizebox{\columnwidth}{!}{
\begin{tabular}{l|c|c|c|c}
                        & \textbf{Pretraining}  & \textbf{Testing}     & \textbf{Overall}       &  \textbf{Proportion of}       \\ 
\textbf{Location}           &  \textbf{(2 hours)}   &  \textbf{(3 hours)}  &  \textbf{(5 hours)}  &  \textbf{the entire dataset}   \\ \hline
Forest                  & 3{,}174                  & 5{,}973                 & 9{,}147                 & 1.27\%    \\
\textcolor{gray}{Clear night} & \textcolor{gray}{845} & \textcolor{gray}{695} & \textcolor{gray}{1{,}540}  &   \\
\textcolor{gray}{Rainy night} & \textcolor{gray}{303} & \textcolor{gray}{477} & \textcolor{gray}{780}   &   \\ 
\textcolor{gray}{Foggy night} & \textcolor{gray}{585} & \textcolor{gray}{602} & \textcolor{gray}{1{,}187}  &   \\
\textcolor{gray}{Clear day}   & \textcolor{gray}{381} & \textcolor{gray}{1{,}467} & \textcolor{gray}{1{,}848}  &   \\
\textcolor{gray}{Rainy day}   & \textcolor{gray}{572} & \textcolor{gray}{1{,}675} & \textcolor{gray}{2{,}247}  &   \\
\textcolor{gray}{Foggy day}   & \textcolor{gray}{488} & \textcolor{gray}{1{,}057} & \textcolor{gray}{1{,}545}  &   \\ \hline
Countryside             & 3{,}525                    & 4{,}283                          & 7{,}808         & 1.09\%  \\ 
\textcolor{gray}{Clear night} & \textcolor{gray}{279} & \textcolor{gray}{1{,}247} & \textcolor{gray}{1{,}526}  &   \\
\textcolor{gray}{Rainy night} & \textcolor{gray}{1{,}137} & \textcolor{gray}{130} & \textcolor{gray}{1{,}267}  &   \\
\textcolor{gray}{Foggy night} & \textcolor{gray}{887} & \textcolor{gray}{795} & \textcolor{gray}{1{,}682}  &   \\
\textcolor{gray}{Clear day}   & \textcolor{gray}{0} & \textcolor{gray}{194} & \textcolor{gray}{194}  &   \\
\textcolor{gray}{Rainy day}   & \textcolor{gray}{1{,}020} & \textcolor{gray}{1{,}312} & \textcolor{gray}{2{,}332}  &   \\
\textcolor{gray}{Foggy day}   & \textcolor{gray}{202} & \textcolor{gray}{605} & \textcolor{gray}{807}  &   \\ \hline
Rural farmland          & 4{,}605                  & 9{,}242                          & 13{,}847          &   1.92\%   \\ 
\textcolor{gray}{Clear night} & \textcolor{gray}{736} & \textcolor{gray}{2{,}631} & \textcolor{gray}{3{,}367}  &   \\
\textcolor{gray}{Rainy night} & \textcolor{gray}{1{,}134} & \textcolor{gray}{265} & \textcolor{gray}{1{,}399}  &   \\
\textcolor{gray}{Foggy night} & \textcolor{gray}{2{,}059} & \textcolor{gray}{2{,}699} & \textcolor{gray}{4{,}758}  &   \\
\textcolor{gray}{Clear day}   & \textcolor{gray}{221} & \textcolor{gray}{1{,}268} & \textcolor{gray}{1{,}489}  &   \\
\textcolor{gray}{Rainy day}   & \textcolor{gray}{418} & \textcolor{gray}{926} & \textcolor{gray}{1{,}344}  &   \\
\textcolor{gray}{Foggy day}   & \textcolor{gray}{37} & \textcolor{gray}{1{,}453} & \textcolor{gray}{1{,}490}  &   \\ \hline
Highway      & 27{,}573                          & 40{,}275                          & 67{,}848            &   9.43\% \\ 
\textcolor{gray}{Clear night} & \textcolor{gray}{4{,}676} & \textcolor{gray}{4{,}878} & \textcolor{gray}{9{,}554}  &   \\
\textcolor{gray}{Rainy night} & \textcolor{gray}{4{,}809} & \textcolor{gray}{4{,}508} & \textcolor{gray}{9{,}317}  &   \\
\textcolor{gray}{Foggy night} & \textcolor{gray}{6{,}052} & \textcolor{gray}{4{,}876} & \textcolor{gray}{10{,}928}  &   \\
\textcolor{gray}{Clear day}   & \textcolor{gray}{3{,}533} & \textcolor{gray}{7{,}575} & \textcolor{gray}{11{,}108}  &   \\
\textcolor{gray}{Rainy day}   & \textcolor{gray}{4{,}235} & \textcolor{gray}{9{,}757} & \textcolor{gray}{13{,}992}  &   \\
\textcolor{gray}{Foggy day}   & \textcolor{gray}{4{,}268} & \textcolor{gray}{8{,}681} & \textcolor{gray}{12{,}949}  &   \\ \hline
Low density   & 56{,}108                          & 84{,}990                         & 141{,}098          &   19.60\%    \\ 
residential   &                                &                               &                 &              \\
\textcolor{gray}{Clear night} & \textcolor{gray}{7{,}348} & \textcolor{gray}{9{,}214} & \textcolor{gray}{16{,}562}  &   \\
\textcolor{gray}{Rainy night} & \textcolor{gray}{6{,}957} & \textcolor{gray}{7{,}381} & \textcolor{gray}{14{,}338}  &   \\
\textcolor{gray}{Foggy night} & \textcolor{gray}{7{,}673} & \textcolor{gray}{8{,}190} & \textcolor{gray}{15{,}863}  &   \\
\textcolor{gray}{Clear day}   & \textcolor{gray}{11{,}486} & \textcolor{gray}{22{,}607} & \textcolor{gray}{34{,}093}   &   \\
\textcolor{gray}{Rainy day}   & \textcolor{gray}{11{,}736} & \textcolor{gray}{17{,}570} & \textcolor{gray}{29{,}306}  &   \\
\textcolor{gray}{Foggy day}   & \textcolor{gray}{10{,}908} & \textcolor{gray}{20{,}028} & \textcolor{gray}{30{,}936}   &   \\\hline
Community      & 23{,}965                          & 42{,}205                         & 66{,}170        &  9.19\%    \\ 
buildings      &                                &                               &              &            \\
\textcolor{gray}{Clear night} & \textcolor{gray}{3{,}648} & \textcolor{gray}{4{,}984} & \textcolor{gray}{8{,}632}   &   \\
\textcolor{gray}{Rainy night} & \textcolor{gray}{3{,}746} & \textcolor{gray}{3{,}532} & \textcolor{gray}{7{,}278}   &   \\
\textcolor{gray}{Foggy night} & \textcolor{gray}{3{,}386} & \textcolor{gray}{4{,}121} & \textcolor{gray}{7{,}507}   &   \\
\textcolor{gray}{Clear day}   & \textcolor{gray}{4{,}838} & \textcolor{gray}{9{,}096} & \textcolor{gray}{13{,}934}  &   \\
\textcolor{gray}{Rainy day}   & \textcolor{gray}{4{,}210} & \textcolor{gray}{9{,}539} & \textcolor{gray}{13{,}749}   &   \\
\textcolor{gray}{Foggy day}   & \textcolor{gray}{4{,}137} & \textcolor{gray}{10{,}933} & \textcolor{gray}{15{,}070}   &   \\ \hline
High density  & 164{,}708                         & 249{,}116                        & 413{,}824       &  57.50\%    \\
residential   &                                &                               &              &             \\
\textcolor{gray}{Clear night} & \textcolor{gray}{26{,}627} & \textcolor{gray}{38{,}134} & \textcolor{gray}{64{,}761}   &   \\
\textcolor{gray}{Rainy night} & \textcolor{gray}{31{,}006} & \textcolor{gray}{25{,}618} & \textcolor{gray}{56{,}624}   &   \\
\textcolor{gray}{Foggy night} & \textcolor{gray}{28{,}064} & \textcolor{gray}{33{,}440} & \textcolor{gray}{61{,}504}   &   \\
\textcolor{gray}{Clear day}   & \textcolor{gray}{26{,}260} & \textcolor{gray}{48{,}795} & \textcolor{gray}{75{,}055}   &   \\
\textcolor{gray}{Rainy day}   & \textcolor{gray}{26{,}830} & \textcolor{gray}{53{,}676} & \textcolor{gray}{80{,}506}   &   \\
\textcolor{gray}{Foggy day}   & \textcolor{gray}{25{,}921} & \textcolor{gray}{49{,}453} & \textcolor{gray}{75{,}374}   &   \\ \hline
\textbf{Total}                     & \textbf{283{,}658}              & \textbf{436{,}084}              & \textbf{719{,}742}   & \textbf{100\%}
\end{tabular}}

\label{tab:Nb_images_dynamic_weather}
\end{table}

\subsection{Analysis of transiting agents}\label{sec:analysis_transient_agent}

\begin{figure*}
  \centering
  \begin{minipage}[b]{0.48\textwidth}
    \centering
    \includegraphics[width=\linewidth]{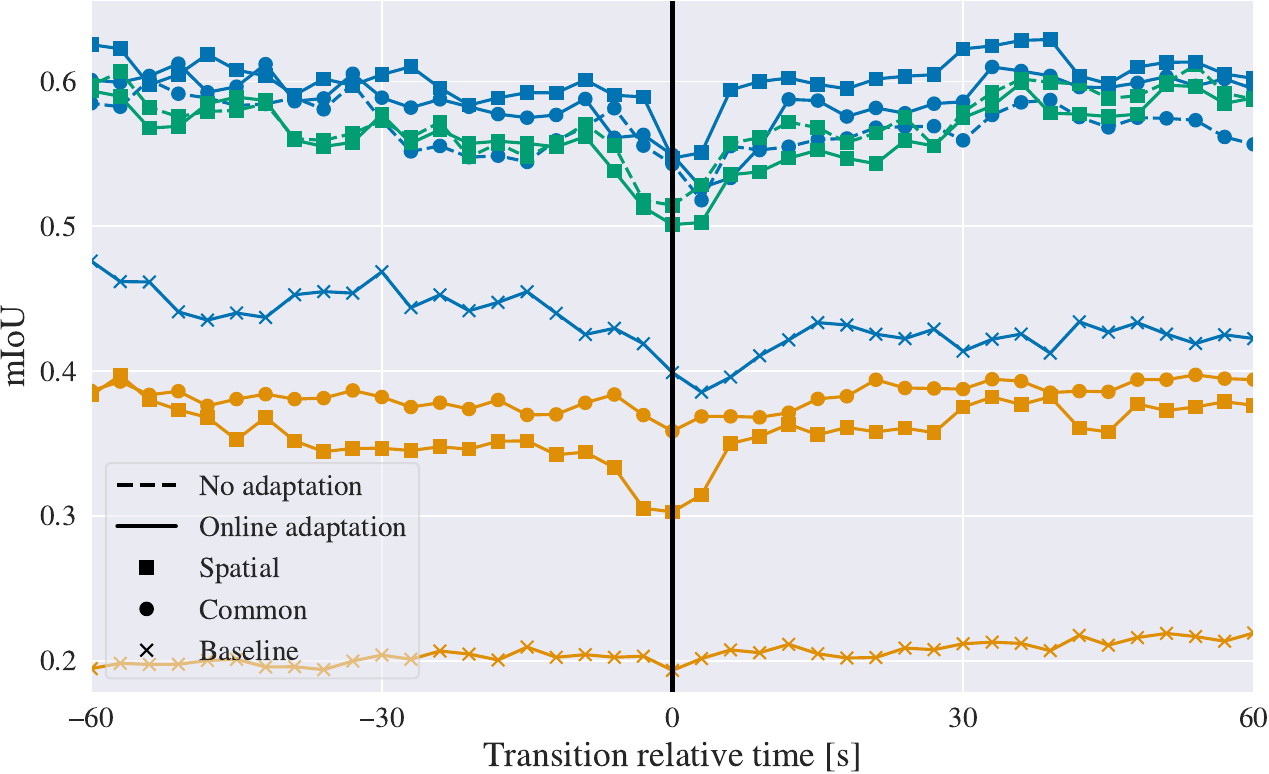}
  \end{minipage}
  \hfill
  \begin{minipage}[b]{0.48\textwidth}
    \centering
    \includegraphics[width=\linewidth]{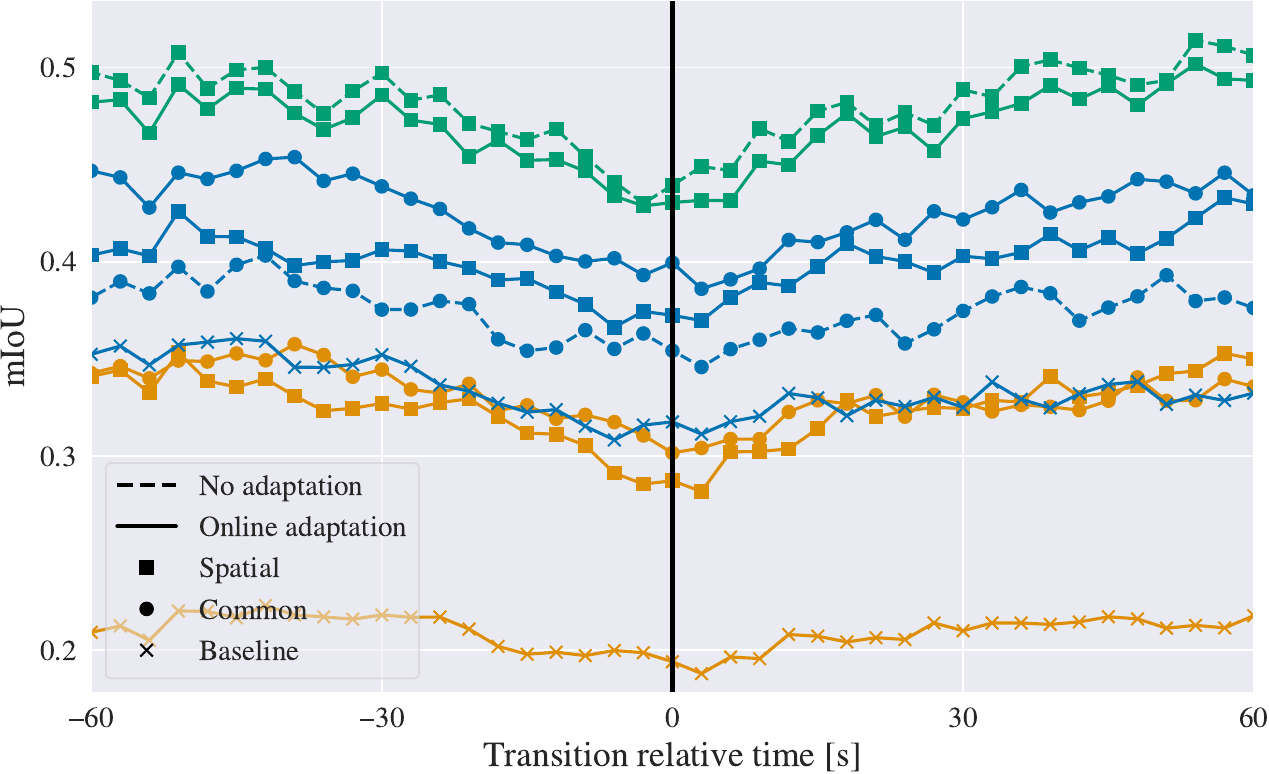}
    
  \end{minipage}
  \centering
  \begin{minipage}[b]{0.48\textwidth}
    \centering
    \includegraphics[width=\linewidth]{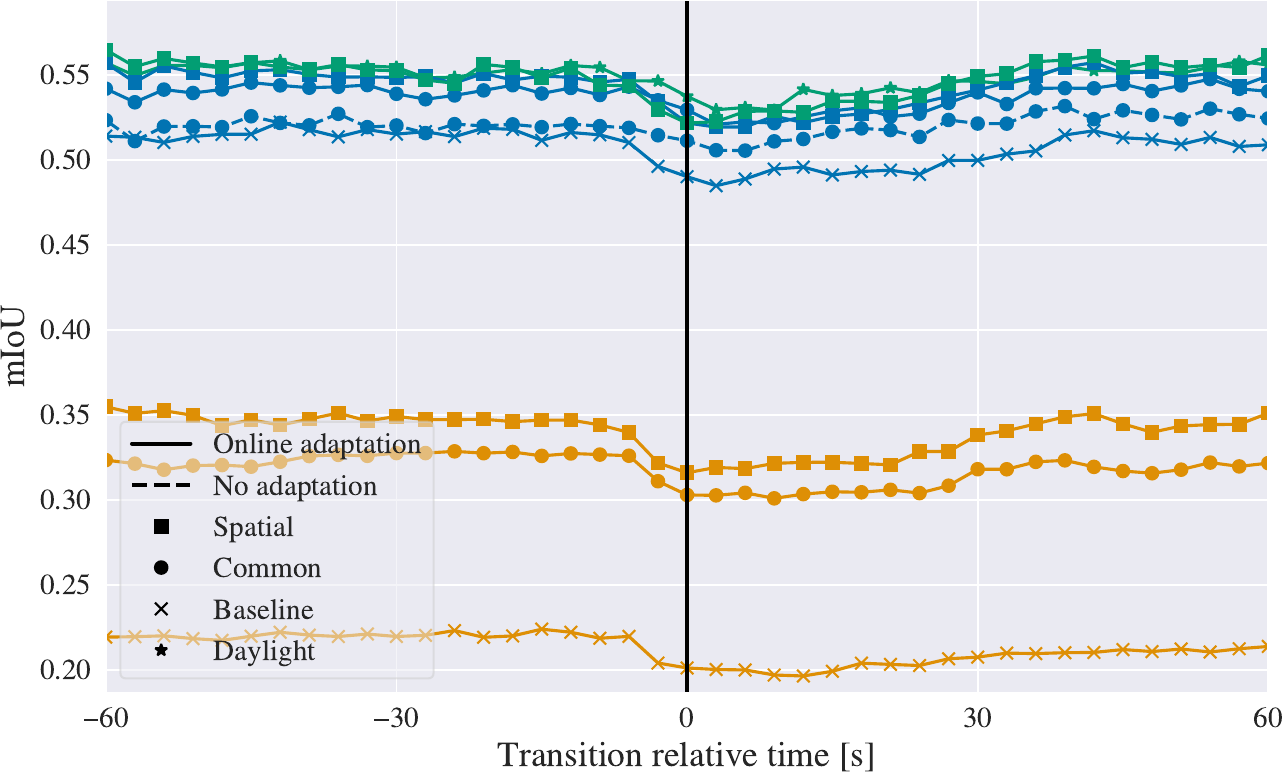}
  \end{minipage}
  \hfill
  \begin{minipage}[b]{0.48\textwidth}
    \centering
    \includegraphics[width=\linewidth]{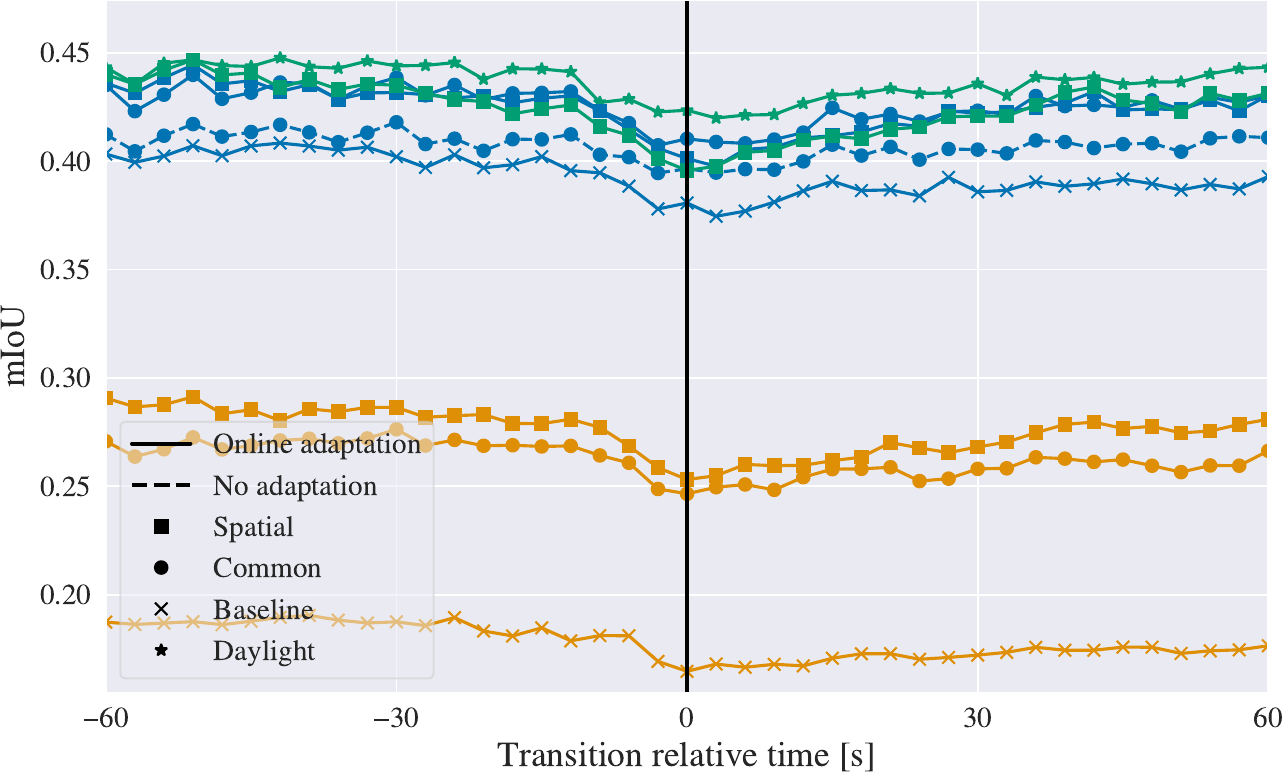}
    
  \end{minipage}
  \caption{\textbf{Fleet performance around cell transitions on DADE-static dataset (top) and DADE-dynamic dataset (bottom).} Comparison of the performance in the MSC-OL setup (left) and the MSC-TTA setup (right) of our method (best settings) with the baseline for each pretraining (\textcolor{Scratch}{\textit{Scratch}}, \textcolor{General}{\textit{General}}, and \textcolor{Cell}{\textit{Cell}}). Confusion matrices of each frame are aggregated using a sliding windows of $3$ seconds. The results are shown $30$ seconds before and after any cell transition that the agents encounter during the $3$ hours of testing.
  } 
    \label{fig:exp_transition} 
\end{figure*}

In this section, we provide insights about the transition of agents between cells. Particularly, we study the evolution of the performance of the models around cell transitions. Figure~\ref{fig:exp_transition} shows the mean performance of the agents transiting from one cell to another, \eg, from a specific location to another, or from a weather condition to another. As can be seen, after the transition, the baseline method experiences a decrease in performance, which remains low for a long period of time. Contrarily, our method is able to recover much faster thanks to the switch between the cell-specific models. 
It is also interesting to observe that for both the MSC-OL and MSC-TTA setups, a temporary drop of mIoU score occurs right before a transition. 
This is probably due to the fact that a vehicle approaching another cell may already see content from an adjacent cell while performing the task with the previous cell's model.
%
%
For instance, a vehicle approaching the city center may record an image with its frontal camera showing the city center while still being registered in another location (\eg, the countryside or the highway). This means that the agent will use the wrong cell model to analyze the environment.
In future work, we aim to address this issue by proposing a model that automatically recognizes the cell, rather than relying on predefined rules.

\subsection{Experiments on cyclic domain shifts}\label{sec:other_datasets}

In Table~\ref{tab:extra_datasets}, we present additional experiments on the dataset and the best method proposed by Houyon~\etal~\cite{Houyon2023Online}, namely \emph{Baseline}+MIR, alongside the performance of the frozen teacher and student trained on the same set (namely, Cityscapes). 
The dataset and method~\cite{Houyon2023Online} are specifically tailored for cyclic domain shifts.
The two first columns are reported from Tables~\ref{tab:agg_sw} and \ref{tab:agg_dw}, for the 3 hours test sets.

Notably, our method demonstrates superior performance on DADE-static, DADE-dynamic, and the cyclic dataset of Houyon~\etal~\cite{Houyon2023Online}.
Furthermore, we see that the \emph{Baseline}+MIR performs worse than the baseline for DADE-static and DADE-dynamic, while it performs better on the cyclic dataset of Houyon~\etal~\cite{Houyon2023Online}. 

\begin{table}[]
    \centering
    \caption{Comparison of our MSC-TTA method with a frozen teacher, a frozen student, the \emph{Baseline}~\cite{Cioppa2019ARTHuS}, and \emph{Baseline}+MIR~\cite{Houyon2023Online}, on our DADE datasets and the dataset of Houyon~\etal~\cite{Houyon2023Online}.}
    \resizebox{\columnwidth}{!}{
    \begin{tabular}{l|c|c|c}
        \textit{mIoU-I}  & DADE-S & DADE-D & Houyon~\cite{Houyon2023Online}\\
        \hline
        \hline
        Teacher~\tiny{\snow} & .668 & .611 & / \\
        Student~\tiny{\snow} & .214 & .159 & / \\
        \hline
        \emph{Baseline}~\cite{Cioppa2019ARTHuS} & .274 & .212 & .234\\
        \emph{Baseline}+MIR~\cite{Houyon2023Online} & .181 & .147 & .256\\
        \textbf{Ours} & \textbf{.362} & \textbf{.312} & \textbf{.277}\\
    \end{tabular}}
    \vspace{-0.4mm}
    \label{tab:extra_datasets}
\end{table}

The results also demonstrate that our method exhibits an expected performance deficit relative to the teacher, while consistently outperforming the student. 
The teacher is a state-of-the-art semantic segmentation model (namely, SegFormer trained on Cityscapes~\cite{Cordts2016The}) and thus exhibits great performance on our DADE datasets. However, the emphasis on achieving the best possible performance often comes with increased complexity and overlooks the critical real-time aspect. The frame rate of SegFormer is approximately $2$ frame per second, it is thus far from being real time. Our proposed method aims at mimicking the performance of available teacher models while reducing computational power and battery usage, thereby bringing state-of-the-art performance at a higher frame rate.

The frozen student is trained on the Cityscapes~\cite{Cordts2016The} dataset with an initial learning rate of $10^{-4}$ using the Adam optimizer for $45$ epochs, reducing the learning rate by a factor of $10$ every $15$ epochs, a cross-entropy loss function, and a batch size of $8$. To match the dimension of the images in the DADE datasets, images from Cityscapes were resized to 720x1440 to keep the same ratio, then cropped to 720x1280.

\end{document}